\documentclass[letterpaper, 10 pt, conference]{cls/ieeeconf}  
\IEEEoverridecommandlockouts                              
\let\labelindent\relax

\usepackage[table,usenames,dvipsnames]{xcolor}
\usepackage{enumitem}
\usepackage[noadjust]{cite}
\usepackage{amsmath,amssymb,amsfonts,amsthm,amscd,dsfont} 
\usepackage{algorithm,algorithmicx,listings}        
\usepackage[noend]{algpseudocode}			        

\usepackage{graphicx,tabularx,adjustbox}
\graphicspath{{./fig/}}
\usepackage{multicol}
\usepackage[font={small}]{caption}   
\usepackage{subcaption}

\usepackage[bookmarks=true, breaklinks=true, colorlinks]{hyperref}
\usepackage{afterpage}

\DeclareMathOperator*{\tr}{tr}

\newtheorem*{proposition*}{Proposition}

\newtheorem*{corollary*}{Corollary}

\theoremstyle{definition}

\newtheorem*{assumption*}{Assumption}
\newtheorem{problem}{Problem}
\newtheorem*{problem*}{Problem}
\theoremstyle{remark}

\newcommand{\calD}{{\cal D}}

\newcommand{\calH}{{\cal H}}

\newcommand{\calJ}{{\cal J}}

\newcommand{\calT}{{\cal T}}

\newcommand{\calV}{{\cal V}}

\newcommand{\calY}{{\cal Y}}

\newcommand{\frakp}{{\mathfrak{p}}}
\newcommand{\frakq}{{\mathfrak{q}}}

\newcommand{\bffrakp}{{\boldsymbol{\mathfrak{p}}}}
\newcommand{\bffrakq}{{\boldsymbol{\mathfrak{q}}}}

\newcommand{\bfd}{\mathbf{d}}
\newcommand{\bfe}{\mathbf{e}}
\newcommand{\bff}{\mathbf{f}}
\newcommand{\bfg}{\mathbf{g}}

\newcommand{\bfp}{\mathbf{p}}

\newcommand{\bfr}{\mathbf{r}}

\newcommand{\bfu}{\mathbf{u}}
\newcommand{\bfv}{\mathbf{v}}

\newcommand{\bfx}{\mathbf{x}}
\newcommand{\bfy}{\mathbf{y}}

\newcommand{\bfzeta}{\boldsymbol{\zeta}}

\newcommand{\bftheta}{\boldsymbol{\theta}}

\newcommand{\bfpi}{\boldsymbol{\pi}}

\newcommand{\bfomega}{\boldsymbol{\omega}}

\newcommand{\bfD}{\mathbf{D}}

\newcommand{\bfI}{\mathbf{I}}
\newcommand{\bfJ}{\mathbf{J}}
\newcommand{\bfK}{\mathbf{K}}
\newcommand{\bfL}{\mathbf{L}}
\newcommand{\bfM}{\mathbf{M}}

\newcommand{\bfQ}{\mathbf{Q}}
\newcommand{\bfR}{\mathbf{R}}

\newcommand{\bfU}{\mathbf{U}}

\newcommand{\bfGamma}{\boldsymbol{\Gamma}}

\newcommand{\bfLambda}{\boldsymbol{\Lambda}}

\newcommand{\bbR}{\mathbb{R}}

\title{\LARGE \bf Hamiltonian Dynamics Learning from Point Cloud Observations for Nonholonomic Mobile Robot Control}

\author{Abdullah Altawaitan \and Jason Stanley \and Sambaran Ghosal \and Thai Duong \and Nikolay Atanasov
\thanks{We gratefully acknowledge support from NSF CCF-2112665 (TILOS).}%
\thanks{The authors are with the Department of Electrical and Computer Engineering, University of California San Diego, La Jolla, CA 92093, USA, e-mails: {\tt\small \{aaltawaitan,\allowbreak jtstanle,\allowbreak sghosal,\allowbreak tduong,\allowbreak natanasov\}@ucsd.edu}. A. Altawaitan is also affiliated with Kuwait University as a holder of a scholarship.}%
}

\begin{document}

\maketitle
\thispagestyle{empty}
\pagestyle{empty}

\begin{abstract}
Reliable autonomous navigation requires adapting the control policy of a mobile robot in response to dynamics changes in different operational conditions. Hand-designed dynamics models may struggle to capture model variations due to a limited set of parameters. Data-driven dynamics learning approaches offer higher model capacity and better generalization but require large amounts of state-labeled data. This paper develops an approach for learning robot dynamics directly from point-cloud observations, removing the need and associated errors of state estimation, while embedding Hamiltonian structure in the dynamics model to improve data efficiency. We design an observation-space loss that relates motion prediction from the dynamics model with motion prediction from point-cloud registration to train a Hamiltonian neural ordinary differential equation. The learned Hamiltonian model enables the design of an energy-shaping model-based tracking controller for rigid-body robots. We demonstrate dynamics learning and tracking control on a real nonholonomic wheeled robot.
\end{abstract}

\section{Introduction}
\label{sec:introduction}

Autonomous mobile robot navigation is playing an increasingly important role in industrial manufacturing, agriculture, transportation, and warehouse automation. Reliable robot use over long duration requires robustness to changing conditions. Autonomous navigation requires adapting robot's control strategy due to its own dynamics which can be done using model-free~\cite{behavioral_systems_theory} or model-based~\cite{mbrl} control techniques.

In this paper, we focus on model-based control of a rigid-body robot for trajectory tracking. The challenge is to infer a robot dynamics model from sensor observations and design a control policy that can successfully execute reference trajectories from a motion planner. Traditionally, a dynamics model is derived from first principles and its parameters are optimized using system identification techniques~\cite{ljung1999system}. Hand-designed models, however, have a limited set of parameters and may not be sufficiently expressive to capture variations in the system dynamics. Recently, machine learning techniques have excelled at learn robot dynamics from data \cite{nguyen2011model,deisenroth2011pilco,williams2017information, raissi2018multistep,chua2018deep} by optimizing a large number of parameters in an expressive model, e.g., neural networks. However, these techniques require large datasets of state trajectories to learn a good dynamics model. 
Physics-informed neural networks reduce the amount of required data by embedding prior knowledge, such as physics laws and kinematic constraints, in the model architecture, ensuring that these requirements are satisfied by design instead of being inferred from data. This leads to better generalization and data efficiency. For example, Lagrangian \cite{roehrl2020modeling, lutter2019deeplagrangian,gupta2019general,cranmer2020lagrangian,lutter2019deepunderactuated} and Hamiltonian formulations \cite{greydanus2019hamiltonian, bertalan2019learning, chen2019symplectic, finzi2020simplifying, Zhong2020Symplectic, willard2020integrating, duong21hamiltonian, havens2021forced, Duruisseaux_LieFVIN_L4DC23} have been used to design neural network models to approximate dynamics of mechanical systems, where the equations of motions are enforced in the neural network architecture. 
Another approach is to assume known nominal dynamics for the system and only learn residual dynamics, e.g., when system dynamics change due to model uncertainty or disturbances. The residual dynamics can be learned online explicitly \cite{mckinnon2021meta, duong2022adaptive, fan2020bayesian, harrison2018control, richards21adaptive, shi2021meta, o2021meta}, allowing adaptation of the controller to dynamics uncertainty, or implicitly by learning controller's parameters \cite{chen2019neural, romero2023MPCC}.
Many physics-informed dynamics learning approaches, however, assume perfect state estimation for supervised training, and hence, are prone to estimation errors or are reliant on motion capture systems, limiting their real-world applications. Instead, our approach learns the robot dynamics directly from sensor observations, while maintaining a Hamiltonian-based machine learning model for data efficiency, generalization, and control design.

\begin{figure}[t]
    \centering
    \includegraphics[width=\linewidth]{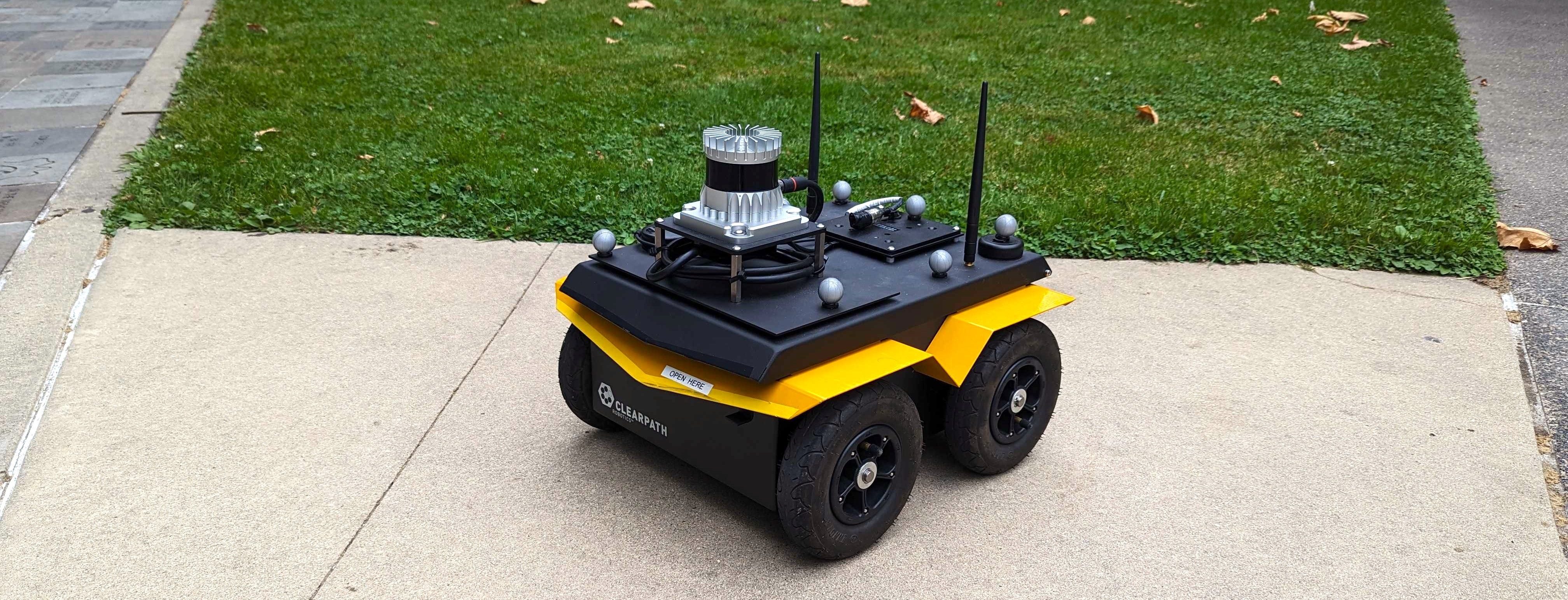}
    \caption{Clearpath Jackal robot equipped with a LiDAR.}
    \label{fig:observations_loss}
    \vspace*{-0.75cm}
\end{figure}

Learning control policies from observations, such as images or point clouds, has been extensively studied in reinforcement learning. A common approach is to learn a latent state space, and train a control policy in the latent state space to maximize the reward or control performance. For example, Zhang et al.~\cite{levine2021learning} learn a latent space for control purposes using bisimulation metrics, which measure the similarity between two latent states if they have similar cumulative future reward. Kamran et al.~\cite{learning_path_tracking} use convolutional neural networks to map images to an embedded space and learn a tracking controller via an actor-critic approach. 
Tian et al.~\cite{latent_model_learning} learn a latent state representation and latent dynamics model from observations to achieve discrete-time linear quadratic Gaussian regulation. The authors establish a finite-sample guarantee for obtaining a near-optimal state representation and control policy. In contrast, our work uses a known state representation but learns non-linear control-affine dynamics with Hamiltonian structure. We use a cycle consistency loss that relates motion prediction from the dynamics model with motion prediction from registration of point-cloud observations. Cycle-consistency has been exploited in computer vision as a self-supervised learning technique to learn a model using unlabeled data for image alignment \cite{Zhou_2015_CVPR, zhou2016learning}, image-to-image translation \cite{zhu2017unpaired}, structure from motion~\cite{zach2010disambiguating,Wilson_2013_ICCV}, and space-time alignment in videos \cite{Dwibedi_2019_CVPR, Wu2021}. Recently, cycle consistency between observations has been used to transfer a learned control policy across different domains \cite{zhang2021learning} without considering robot dynamics. 

In this work, we develop a cycle consistency approach for robot dynamics learning from sensor observations. Our approach predicts the robot motion using a Hamiltonian dynamics model and measures the motion error using point cloud registration. Our prior work \cite{duong21hamiltonian, duong2024port} developed a Hamiltonian neural ODE (HNODE) method for learning quadrotor dynamics and an energy-shaping control formulation for quadrotor trajectory tracking. In this paper, we apply the HNODE approach to learn nonholonomic dynamics of a ground robot and supervise the learning process directly from sensor observations.
Given a control input, we predict the robot's next pose using the dynamics model and obtain an observation from the sensors. The observations at two consecutive steps should be consistent, e.g., features from the first observation, after transformation to the next frame, should appear in the second observation. 
Violations of this cycle-consistency principle create a loss function that we use to train the Hamiltonian dynamics model via a neural ODE framework~\cite{chen2018neural}. 
We then develop an energy-shaping trajectory tracking controller for the learned~dynamics and demonstrate our approach in simulated and real autonomous navigation experiments with nonholonomic wheeled robots.

\section{Problem Statement}
\label{sec:problem_statement}

Consider a robot modeled as a single rigid body with pose represented by position $\bfp \in \bbR^3$ and rotation matrix $\bfR \in SO(3)$. Let $\bfr_1, \bfr_2, \bfr_3 \in \bbR^3$ be the rows of $\bfR$, and define the robot's generalized coordinates as $\boldsymbol{\mathfrak{q}} = \begin{bmatrix} \bfp^{\top} & \bfr_1^{\top} & \bfr_2^{\top} & \bfr_3^{\top} \end{bmatrix}^{\top} \in \mathbb{R}^{12}$. Denote the generalized velocity of the robot by $\bfzeta = \begin{bmatrix} \bfv^{\top} & \bfomega^{\top} \end{bmatrix}^{\top} \in \mathbb{R}^6$, where $\bfv \in \mathbb{R}^3$ is the linear velocity and $\bfomega \in \mathbb{R}^3$ is the angular velocity, expressed in the body frame. The evolution of the robot state $\bfx = \begin{bmatrix} \boldsymbol{\mathfrak{q}}^{\top} & \bfzeta^{\top} \end{bmatrix}^{\top} \in \mathbb{R}^{18}$ is governed by control input $\bfu$ according to a continuous-time dynamics model:
\begin{equation} \label{eq:general_dynamics}
\dot{\bfx}(t) = \bff(\bfx(t), \bfu(t)).
\end{equation}
We assume that the robot is equipped with a distance sensor, such as depth camera or LiDAR, which provides point clouds $\calY_n = \{\bfy_{n,m} \in \bbR^3\}_{m=1}^M$ at discrete time steps $t_n$. Assuming that the robot dynamics model is unknown, our objective is to design a control policy using a dataset of control inputs and point-cloud observations such that the robot is able to track a desired reference trajectory.

\begin{problem}\label{problem}
Consider a rigid-body robot with unknown dynamics model in \eqref{eq:general_dynamics}. Let $\calD = \{t_n^{(i)}, \calY_n^{(i)}, \bfx_0^{(i)}, \bfu_n^{(i)}\}_{i=1,n=0}^{D,N}$ be a training dataset of $D$ trajectories. Each trajectory $i$ consists of an initial state $\bfx_0^{(i)}$, control inputs $\bfu_n^{(i)}$ applied at times $t_n^{(i)}$ and held constant during $[t_n^{(i)},t_{n+1}^{(i)})$, and point clouds $\calY_n^{(i)}$ received by the robot at $t_n^{(i)}$ for $n = 0,\ldots,N$. Design a control policy $\bfu = \bfpi(\bfx, \bfx^*(t))$ using the data $\calD$ such that the robot is able to track a desired reference trajectory $\bfx^*(t) = [\bffrakq^*(t)^\top\;\bfzeta^*(t)^\top]^\top$ for $t \in [0,\infty)$. 
\end{problem}

\section{Learning From Observations Using Cycle Consistency}
\label{sec:learning_using_cycle_consistency}

We develop a model-based approach to solve Problem~\ref{problem}. We first learn a model of the system from point-cloud observations and then design a control policy $\bfpi(\bfx, \bfx^*(t))$ based on the learned model. We use a deep neural network model $\bff_{\bftheta}$ to approximate the unknown dynamics $\bff$ in \eqref{eq:general_dynamics}. Training a continuous-time neural ODE model can be done via the adjoint method \cite{chen2018neural}. Instead of a dataset $\calD$ of observations and controls as in Problem~\ref{problem}, the usual formulation uses a dataset $\{t_n^{(i)}, \bfx_n^{(i)}, \bfu_n^{(i)}\}_{i=1,n=0}^{D,N}$ of states and controls. The parameters $\bftheta$ are determined by minimizing a loss function, measuring the distance between predicted and actual states:
\begin{align} \label{eq:neural_ode_learning_problem}
    & \min_{\bftheta}
    && \sum_{i=1}^{D} \sum_{n=1}^{N}  \ell_s(\bfx_n^{(i)}, \tilde{\bfx}_n^{(i)})  + \ell_r(\bftheta)\\
    & \text{s.t.} 
    && \dot{\tilde{{\bf x}}}^{(i)}(t) = {\bf f}_{\boldsymbol{\theta}} ({\tilde{{\bf x}}}^{(i)}(t),  {\bf u}^{(i)}), \quad {\tilde{{\bf x}}}^{(i)}(t_0) = {\bf x}^{(i)}_0  \notag\\ 
    &&& \tilde{\bf x}^{(i)}_n = \tilde{\bf x}^{(i)}(t_n), \quad \forall n = 1, \ldots, N, \quad \forall i = 1, \ldots, D, \notag
\end{align} 
with the predicted states $\tilde{\bfx}_n^{(i)}$ obtained from an ODE solver:
\begin{equation*}
\tilde{\bfx}_{n}^{(i)} = \text{ODESolver}(\bfx_0^{(i)}, \bff_{\bftheta}, \bfu_0^{(i)}, \ldots, \bfu_n^{(i)}, t_0^{(i)}, \ldots, t_n^{(i)}).
\end{equation*}
The error $\ell_r$ is a regularization term for the network parameters $\bftheta$, while the error $\ell_s$ can be defined as a sum of square distance functions in $\bbR^3$ for position, $SO(3)$ for orientation, and $\bbR^6$ for generalized velocity (e.g., see \cite{duong21hamiltonian}). The gradient of the loss function with respect to $\bftheta$ is obtained by solving an adjoint ODE backwards in time \cite{chen2018neural}. In this work, we extend the neural ODE formulation by (i) choosing $\bff_{\boldsymbol{\theta}}$ to satisfy Hamilton's equations of motion, which guarantee energy conservation and kinematic constraints, and (ii) training the model parameters $\bftheta$ using point-cloud observations $\calY_n^{(i)}$ rather than system states $\bfx_n^{(i)}$.

\subsection{Imposing Hamiltonian Dynamics}
\label{subsec: Hamiltonian}
To ensure energy conservation, the structure of $\bff_{\bftheta}$ can be designed to follow Hamiltonian dynamics \cite{Zhong2020Symplectic,duong21hamiltonian}:
\begin{equation}\label{eq:port_hamiltonian_form}
\begin{bmatrix}
\dot{\bffrakq}\\\dot{\bffrakp}
\end{bmatrix} = \calJ_{\bftheta}(\bffrakq,\bffrakp) \begin{bmatrix}
            \nabla_{\bffrakq} \calH_{\bftheta}(\bffrakq,\bffrakp) \\ 
            \nabla_{\bffrakp} \calH_{\bftheta}(\bffrakq,\bffrakp)
        \end{bmatrix} + \begin{bmatrix}
            \bf0 \\ \bfg_{\bftheta}(\bffrakq)
        \end{bmatrix} \bfu,
\end{equation}
where $\bffrakp = \bfM_{\bftheta}(\bffrakq) \bfzeta$ is the system momentum, $\bfM_{\bftheta}(\bffrakq)$ is a positive semidefinite generalized mass matrix, $\bfg_{\bftheta}(\bffrakq)$ is an input gain matrix, $\calJ_{\bftheta}(\bffrakq,\bffrakp)$ is a state interconnection matrix, and $\calH_{\bftheta}(\bffrakq,\bffrakp)$ is the Hamiltonian modeling the total energy of the system as a sum of kinetic energy $\calT_{\bftheta}(\bffrakq, \bffrakp) = \frac{1}{2} \bffrakp^{\top} \bfM_{\bftheta}(\bffrakq)^{-1} \bffrakp$ and potential energy $\calV_{\bftheta}(\bffrakq)$:
\begin{equation}\label{eq:hamiltonian_function}
\calH_{\bftheta}(\bffrakq,\bffrakp) = \calT_{\bftheta}(\bffrakq, \bffrakp) + \calV_{\bftheta}(\bffrakq).
\end{equation} 
The dynamics of the generalized velocity $\bfzeta$ are described by $\dot{\bfzeta} = \dot{\bfM}_{\bftheta}(\boldsymbol{\frakq}) \boldsymbol{\frakp} + \bfM_{\bftheta}(\boldsymbol{\frakq}) \dot{\boldsymbol{\frakp}}$.
Eq. \eqref{eq:port_hamiltonian_form} and $\dot{\bfzeta}$ describe our approximated dynamics function $\bff_{\bftheta}(\bfx, \bfu)$ in \eqref{eq:neural_ode_learning_problem}.
To model energy dissipation (e.g., friction) and ensure the kinematic constraints of the orientation matrix $\bfR$, the matrix $\calJ_{\bftheta}(\bffrakq,\bffrakp)$ in \eqref{eq:port_hamiltonian_form} needs to have the following structure \cite{duong21hamiltonian}:
\begin{equation}
\calJ_{\bftheta}(\bffrakq,\bffrakp) = \begin{bmatrix}
       \mathbf{0} & \bffrakq^{\times} \\ 
       -\bffrakq^{\times \top} & \bffrakp^{\times} \end{bmatrix} - \begin{bmatrix} \mathbf{0} & \mathbf{0}\\ \mathbf{0} & \bfD_{\bftheta}(\bffrakq,\bffrakp)
       \end{bmatrix},
\end{equation}
where the first matrix is skew-symmetric and the second matrix models energy dissipation with positive semidefinite $\bfD_{\bftheta}(\bffrakq,\bffrakp)$. The operators, $\bffrakq^{\times}$ and $\bffrakp^{\times}$, are defined as:
\begin{equation}\label{eq:operators_definition}
       \bffrakq^{\times} = 
       \begin{bmatrix}
           \bfR^{\top} & \mathbf{0} & \mathbf{0} & \mathbf{0} \\ 
           \mathbf{0} & \hat{\bfr}_1^{\top} & \hat{\bfr}_2^{\top} & \hat{\bfr}_3^{\top}
       \end{bmatrix}^{\top}, \quad \bffrakp^{\times} = 
       \begin{bmatrix}
           \mathbf{0} & \hat{\bffrakp}_{\bfv} \\
           \hat{\bffrakp}_{\bfv} & \hat{\bffrakp}_{\bfomega}
       \end{bmatrix},
\end{equation}
where $\hat{(\cdot)}: \mathbb{R}^n \mapsto \mathfrak{so}(n)$ maps $\bfx \in \mathbb{R}^n$ to a skew-symmetric matrix $\hat{\bfx} \in \mathfrak{so}(n)$. This model is referred to as HNODE with $\bfM_{\bftheta}(\frakq)$, $\calV_{\bftheta}(\bffrakq)$, $\bfg_{\bftheta}(\bffrakq)$, $\bfD_{\bftheta}(\bffrakq,\bffrakp)$ as neural networks. Instead, we introduce a nominal model in each of the learnable terms and only learn residual components as follows:
\begin{equation}\label{eq:residuals}
\begin{aligned}
&\bfM_{\bftheta}(\bffrakq) = (\bar{\bfL}(\bffrakq) + \tilde{\bfL}_{\bftheta}(\bffrakq))(\bar{\bfL}(\bffrakq) + \tilde{\bfL}_{\bftheta}(\bffrakq))^\top,\\
&\bfD_{\bftheta}(\bffrakq,\bffrakp) = (\bar{\bfLambda}(\bffrakq) + \tilde{\bfLambda}_{\bftheta}(\bffrakq))(\bar{\bfLambda}(\bffrakq) + \tilde{\bfLambda}_{\bftheta}(\bffrakq))^\top,\\
&\calV_{\bftheta}(\bffrakq) = \bar{\calV}(\bffrakq) + \tilde{\calV}_{\bftheta}(\bffrakq),
\quad \bfg_{\bftheta}(\bffrakq) = \bar{\bfg}(\bffrakq) + \tilde{\bfg}_{\bftheta}(\bffrakq),
\end{aligned}
\end{equation}
where the terms with bar and tilde accents represent fixed nominal models and learnable residuals, respectively. The structure of $\bfM_{\bftheta}(\bffrakq)$ and $\bfD_{\bftheta}(\bffrakq,\bffrakp)$ is chosen to ensure that they remain positive semidefinite using lower-triangular matrices $\bar{\bfL}(\bffrakq)$, $\tilde{\bfL}_{\bftheta}(\bffrakq)$, $\bar{\bfLambda}(\bffrakq)$, $\tilde{\bfLambda}_{\bftheta}(\bffrakq)$.

\subsection{Training from Observations}

Next, we focus on reformulating dynamics learning problem in \eqref{eq:neural_ode_learning_problem} to use point-cloud observations $\calY_n^{(i)}$ instead of states $\bfx_n^{(i)}$. While point-cloud observations are available directly from RGBD cameras and LiDAR sensors found on most mobile robots, the robot states need to be estimated via LiDAR-based odometry, visual-inertial odometry, or a motion capture system. State estimation introduces localization errors in the dynamics learning process or requires additional infrastructure in the case of motion capture.

To enable learning from point-cloud observations, we introduce an observation-space error function $\ell_o$ and observation predictions $\tilde{\bfy}_{n,m}^{(i)}$ in the dynamics learning problem:
\begin{align}\label{eq:learning_problem}
    & \min_{\bftheta}
    && \sum_{i=1}^{D} \sum_{n=1}^{N} \sum_{m=1}^{M} \ell_o(\bfy_{n,m}^{(i)}, \tilde{\bfy}_{n,m}^{(i)}) + \ell_{r}({\bftheta})   \notag\\
    & \text{s.t.} 
    && \tilde{\bfy}_{n+1,m}^{(i)} = \tilde{\bfR}_{n+1}^{(i)\top} (\tilde{\bfR}_{n}^{(i)} \tilde{\bfy}_{n,m}^{(i)} + \tilde{\bfp}_{n}^{(i)} - \tilde{\bfp}_{n+1}^{(i)}), \notag\\ 
    &&& \dot{\tilde{{\bf x}}}^{(i)}(t) = {\bf f}_{\boldsymbol{\theta}} ({\tilde{{\bf x}}}^{(i)}(t),  {\bf u}^{(i)}), \\ 
    &&& {\tilde{{\bf x}}}^{(i)}(t_0) = {\bf x}^{(i)}_0, \quad \tilde{\bfy}_{0,m}^{(i)} = \bfy_{0,m}^{(i)}, \notag\\
    &&& \tilde{\bf x}^{(i)}_n = \tilde{\bf x}^{(i)}(t_n), \quad \forall n = 0, \ldots, N, \quad \forall i = 1, \ldots, D,\notag
\end{align}
where $\ell_o(\bfy_{n,m}^{(i)}, \tilde{\bfy}_{n,m}^{(i)})$ can be selected as the square Euclidean distance $\| \bfy_{n,m}^{(i)} - \tilde{\bfy}_{n,m}^{(i)} \|_2^{2}$.  The correspondence $(\bfy_{n,m}^{(i)}, \bfy_{n+1,m}^{(i)})$ is obtained using point-cloud registration~\cite{zhou2016fast}. The motivation for using a loss function in the observation space, instead of recovering the robot pose first, is to enable more general robot observations (e.g, image features) and states (e.g., velocity) in future work. The loss function can be modified to handle outliers, e.g., via a robust estimator, and even consider data association optimization.
The key idea in this formulation is to relate motion predictions $\tilde{\bfx}_{n+1}^{(i)}$ from the learned dynamics with point-cloud prediction $\tilde{\bfy}_{n+1,m}^{(i)}$ in the first constraint in \eqref{eq:learning_problem}. As illustrated in Fig.~\ref{fig:observations_loss}, a point cloud $\calY_{n}^{(i)}$ observed at time $t_n$ in the robot's body-frame is transformed to the world-frame using the predicted state $\tilde{\bfx}_{n}^{(i)}$. With the control input $\bfu_n^{(i)}$ at $t_n$, the model predicts the next state $\tilde{\bfx}_{n+1}^{(i)}$ at $t_{n+1}$, and transforms the 3D points back to the robot's body frame to obtain $\tilde{\calY}_{n+1}^{(i)}$. The discrepancy between these predictions and the actual observations is captured by $\ell_o$. The term $\ell_r$ in \eqref{eq:learning_problem} is used to promote sparsity in the learned dynamics model. For example, in underactuated systems the input gain $\bfg_{\bftheta}$ is often sparse because some inputs affect a subset of the states. Promoting sparsity through regularization ensures that the model remains simple and reflects the underactuation model structure accurately.

\begin{figure}[t]
    \centering
    \includegraphics[width=\linewidth,%
                           trim=0mm 25mm 0mm 5mm,clip]{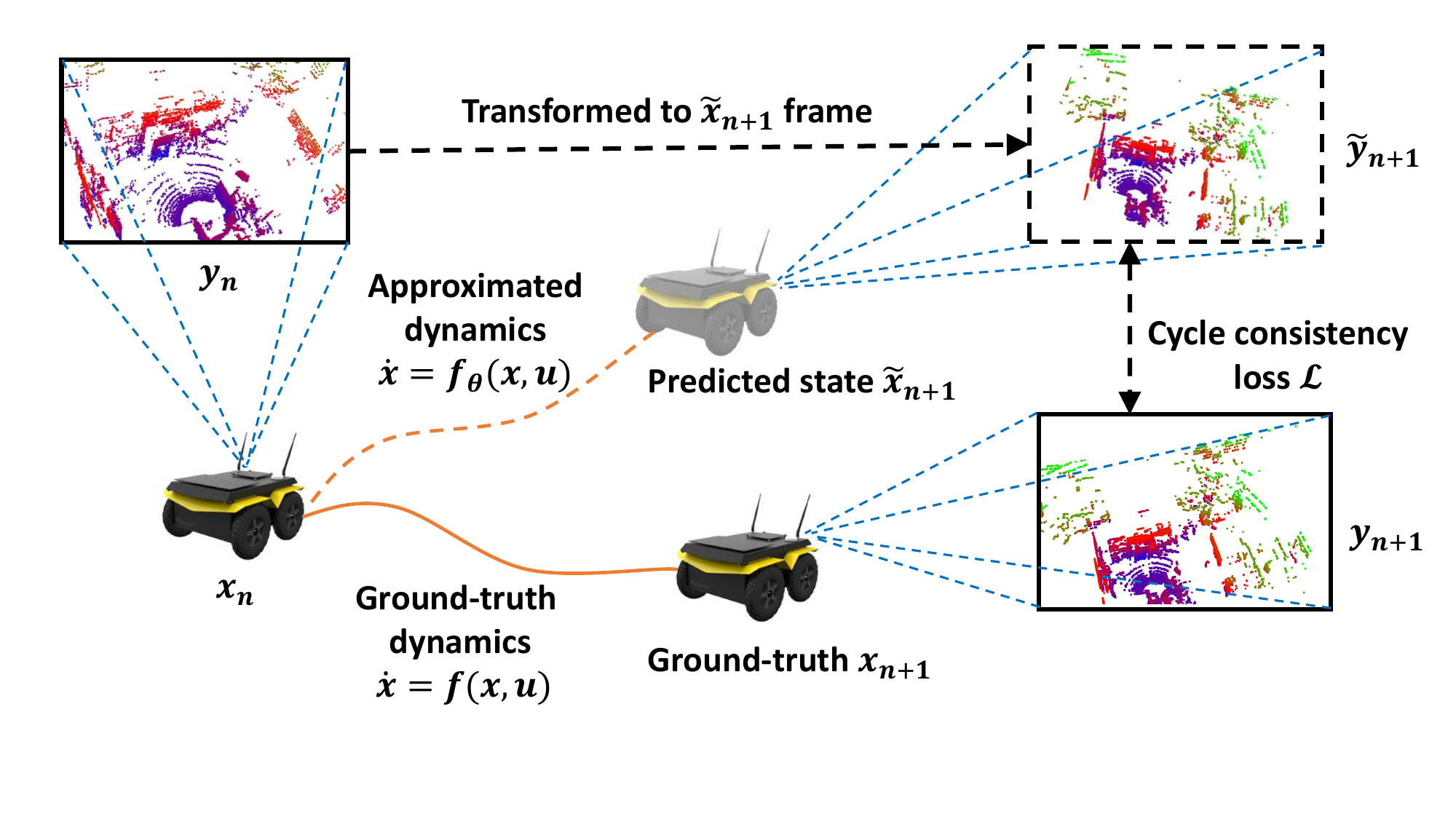}
    \caption{Illustration of our observation-space loss function design.}
    \label{fig:observations_loss}
    \vspace*{-0.5cm}
\end{figure}

\section{Control Design For Trajectory Tracking}
\label{sec:control_design}

Given the Hamiltonian dynamics model learned from observations in Sec. \ref{sec:learning_using_cycle_consistency}, a tracking control policy is designed using the interconnection and damping assignment passivity-based control (IDA-PBC) approach \cite{van2014port, wang2008modified, souza2014passivity}. We present a general approach in Sec.~\ref{subsec:ida_pbc_approach} and apply it to nonholonomic wheeled robots in Sec.~\ref{subsec:ida_pbc_differential_drive}.

\subsection{IDA-PBC Control Approach} \label{subsec:ida_pbc_approach}
Let $\bfx^*(t) = [\bffrakq^*(t)^\top\;\bfzeta^*(t)^\top]^\top$ be the desired trajectory that the system should track. For readability, we omit the time dependence of $\bfx^*$ in the remainder of the paper. The desired momentum of the system in the body frame is calculated as $\bffrakp^* = \bfM \begin{bmatrix} \bfR^\top \bfR^*\bfv^* \\ \bfR^\top \bfR^* \bfomega^*\end{bmatrix}$, leading to the desired $(\boldsymbol{\mathfrak{q}}^{*}, \boldsymbol{\mathfrak{p}}^{*})$ for the learned Hamiltonian system.
The Hamiltonian function $\calH(\bffrakq, \bffrakp)$ is generally not minimized at $(\bffrakq^*, \bffrakp^*)$. Therefore, the key idea of IDA-PBC is to inject energy to the system via the control input to achieve a desired Hamiltonian $\calH_d(\bffrakq, \bffrakp)$ that has a minimum at $(\bffrakq^*, \bffrakp^*)$. 
Let $(\bffrakq_e, \bffrakp_e)$ be the error in generalized coordinates and momentum:
\begin{equation}\label{eq:error_state}
\begin{aligned}
\bfR_e& = \bfR^{*\top} \bfR =  \begin{bmatrix} \bfr_{e1} & \bfr_{e2} & \bfr_{e3} \end{bmatrix}^\top, &\bfp_e &= \bfp - \bfp^*,\\
\bffrakq_e &= \begin{bmatrix} \bfp_e^\top & \bfr_{e1}^\top & \bfr_{e2}^\top & \bfr_{e3}^\top \end{bmatrix}^\top, &\bffrakp_e& = \bffrakp-\bffrakp^*.
\end{aligned}
\end{equation}
A common choice of $\calH_d(\bffrakq, \bffrakp)$ is:
\begin{align}
\calH_d(\bffrakq, \bffrakp) &=  \calT_d(\bffrakq_e,\bffrakp_e) + \calV_d(\bffrakq_e), \label{eq:desired_hamiltonian}
\end{align}
where $\calT_d(\bffrakq_e,\bffrakp_e)$ and $ \calV_d(\bffrakq_e)$ are the desired kinetic and potential energy, respectively. 

The IDA-PBC designs a control policy $\bfpi(\bfx, \bfx^*)$ such that the dynamics of the closed-loop system are also Hamiltonian with $\calH_d(\bffrakq, \bffrakp)$ as the total energy:
\begin{equation} \label{eq:closed_loop_error_dynamics}
\begin{bmatrix}
\dot{\bffrakq}_e \\
\dot{\bffrakp}_e \\
\end{bmatrix}
= \mathcal{J}_d(\mathbf\bffrakq_e, \mathbf\bffrakp_e)
\begin{bmatrix}
\frac{\partial \mathcal{H}_d}{\partial \mathbf\bffrakq_e} \\
\frac{\partial \mathcal{H}_d}{\partial \mathbf\bffrakp_e} 
\end{bmatrix},
\end{equation}
where $\mathcal{J}_d(\mathbf\bffrakq_e, \mathbf\bffrakp_e)$ is chosen in our design as follows \cite{duong21hamiltonian}:
\begin{equation}\label{eq:Jd_Rd_design}
\begin{aligned}
&\mathcal{J}_d(\mathbf\bffrakq_e, \mathbf\bffrakp_e) = \begin{bmatrix}
\bf0 & \bfJ_1 \\
-\bfJ_1^\top & \bfJ_2\end{bmatrix} - \begin{bmatrix}
\bf0 & \bf0 \\
\bf0 & \bfK_\bfd 
\end{bmatrix},\\
&\bfJ_1 = \begin{bmatrix}
        \bfR^{\top} & \bf0 & \bf0 & \bf0 \\ 
        \bf0 & \hat{\bfr}_{e1}^{\top} & \hat{\bfr}_{e2}^{\top} & \hat{\bfr}_{e3}^{\top}
    \end{bmatrix}^{\top}, \; 
    \bfJ_2 = \bf0.
\end{aligned}
\end{equation}
By matching the original dynamics in \eqref{eq:port_hamiltonian_form} and the closed-loop dynamics \eqref{eq:closed_loop_error_dynamics}, we observe that the control input $\bfu$ has to satisfy the following matching equations:
\begin{align}
    {\bf 0} &= {\bf J}_1 \frac{\partial \mathcal{H}_d}{\partial \boldsymbol{\mathfrak{p}}_{e}} - \boldsymbol{\mathfrak{q}}^{\times} \frac{\partial \mathcal{H}_{\bftheta}}{\partial \boldsymbol{\mathfrak{p}}} + \dot{\boldsymbol{\mathfrak{q}}} - \dot{\boldsymbol{\mathfrak{q}}}_{e}, \label{eq:matching_equations_kinematics}\\ 
     {\bf g}(\boldsymbol{\mathfrak{q}}) {\bf u} &= \boldsymbol{\mathfrak{q}}^{\times \top} \frac{\partial \mathcal{H}_{\bftheta}}{\partial \boldsymbol{\mathfrak{q}}} - {\bf J}^{\top}_1 \frac{\mathcal{H}_d}{\partial \boldsymbol{\mathfrak{q}}_e} + {\bf J}_2 \frac{\partial \mathcal{H}_d}{\partial \boldsymbol{\mathfrak{p}}_e} - \boldsymbol{\mathfrak{p}}^{\times} \frac{\partial \mathcal{H}_{\bftheta}}{\partial \boldsymbol{\mathfrak{p}}} \notag \\
    & \phantom{{}={}} - {\bf K}_{\bfd} \frac{\partial \mathcal{H}_d}{\partial \boldsymbol{\mathfrak{p}}_e} + \bfD_{\bftheta}(\boldsymbol{\bffrakq}, \boldsymbol{\bffrakp}) \frac{\partial \mathcal{H}_{\bftheta}}{\partial \boldsymbol{\mathfrak{p}}}  + \dot{\boldsymbol{\mathfrak{p}}} - \dot{\boldsymbol{\mathfrak{p}}}_e.
    \label{eq:matching_equations_momentum}
\end{align}
Our choice of $\bfJ_1$ and $\bfJ_2$ in \eqref{eq:Jd_Rd_design} satisfies the condition \eqref{eq:matching_equations_kinematics}. We obtain the control policy $\bfu = \bfpi(\bfx, \bfx^*)$ from \eqref{eq:matching_equations_momentum} as the sum $\bfu = \bfu_{ES} + \bfu_{DI}$ of an energy-shaping component $\bfu_{ES}$ and a damping-injection component $\bfu_{DI}$:
\begin{align} \label{eq:u_ES_DI}
    \bfu_{ES} =& \bfg^{\dagger}(\mathbf\bffrakq)\left(\mathbf\bffrakq^{\times\top}\frac{\partial \calH_{\bftheta}}{\partial \mathbf\bffrakq} - \mathbf\bfJ_1^{\top}\frac{\partial \calH_d}{\partial \mathbf\bffrakq_e} + \bfJ_2\frac{\partial \calH_d}{\partial \mathbf\bffrakp_e} \right. \notag \\
& \left.  \qquad- \mathbf\bffrakp^{\times}\frac{\partial \calH_{\bftheta}}{\partial \mathbf\bffrakp} + \bfD_{\bftheta}(\bffrakq, \bffrakp)\frac{\partial \calH_{\bftheta}}{\partial \mathbf\bffrakp} + \dot{\mathbf\bffrakp} - \dot{\mathbf\bffrakp}_e \right), \notag\\
\bfu_{DI} =& -\bfg^{\dagger}(\mathbf\bffrakq)\bfK_\bfd\frac{\partial \calH_d}{\partial \mathbf\bffrakp_e},
\end{align}
where $\bfg^{\dagger}(\mathbf\bffrakq) = \left(\bfg^{\top}(\mathbf\bffrakq)\bfg(\mathbf\bffrakq)\right)^{-1}\bfg^{\top}(\mathbf\bffrakq)$ is the pseudo-inverse of $\bfg(\mathbf\bffrakq)$. Let $\bfg^{\perp}(\bffrakq)$ be a maximal-rank left annihilator of $\bfg(\bffrakq)$, i.e., $\bfg^{\perp}(\bffrakq)\bfg(\bffrakq) = \bf0$. The control policy \eqref{eq:u_ES_DI} exists when the following matching condition is satisfied:
\begin{align}\label{eq:matching_condition_existence}
    \bfg^{\perp}(\bffrakq) \left(\boldsymbol{\mathfrak{q}}^{\times \top} \frac{\partial \mathcal{H}_{\bftheta}}{\partial \boldsymbol{\mathfrak{q}}} - {\bf J}^{\top}_1 \frac{\mathcal{H}_d}{\partial \boldsymbol{\mathfrak{q}}_e} + {\bf J}_2 \frac{\partial \mathcal{H}_d}{\partial \boldsymbol{\mathfrak{p}}_e} - \boldsymbol{\mathfrak{p}}^{\times} \frac{\partial \mathcal{H}_{\bftheta}}{\partial \boldsymbol{\mathfrak{p}}} \right. &\\
    \left.\phantom{{}={}} - {\bf K}_{\bfd} \frac{\partial \mathcal{H}_d}{\partial \boldsymbol{\mathfrak{p}}_e} + \bfD(\boldsymbol{\bffrakq}, \boldsymbol{\bffrakp}) \frac{\partial \mathcal{H}_{\bftheta}}{\partial \boldsymbol{\mathfrak{p}}}  + \dot{\boldsymbol{\mathfrak{p}}} - \dot{\boldsymbol{\mathfrak{p}}}_e\right) &=0. \notag
\end{align}

\subsection{Control Design for Nonholonomic Wheeled Robots} \label{subsec:ida_pbc_differential_drive}

In this section, we consider an IDA-PBC control policy design for a nonholonomic wheeled robot. The wheels of a nonholonomic rigid-body only allow forces to be applied along the body's $x$-axis and torques around the $z$-axis. With two control inputs, we are able to control at most two degrees of freedom in the robot's configuration $\bffrakq$, making the system underactuated. The desired closed-loop dynamics \eqref{eq:closed_loop_error_dynamics} need to be chosen carefully to ensure that there exists a control policy that can achieve them. Inspired by Lyapunov function designs in prior works \cite{aicardi95, Lages2019}, where a nonholonomic wheeled robot is steered to approach a target pose at a specified angle, we design a desired Hamiltonian function that accounts for the underactuation. A natural choice of the desired kinetic and potential energies for a fully-actuated rigid-body robot is $\calT_d(\bffrakq_e,\bffrakp_e) = \frac{1}{2} \bffrakp_e^\top \bfM(\bffrakq_e)^{-1} \bffrakp_e$ and $\calV_d(\bffrakq_e) = \frac{k_{\bfp}}{2} \bfp_{e}^{\top} \bfp_{e} + \frac{k_{\bfR}}{2} \tr\big( \bfI - \bfR_e \big)$,
where $\calV_d(\bffrakq_e)$ measures the quadratic Euclidean and chordal distance \cite{rotation_averaging}, respectively, from the current to the desired position and orientation with $k_{\bfp}$ and $k_{\bfR}$ as positive scalars. In the case of a nonholonomic wheeled robot, the orientation part of the desired potential energy needs to be shaped to guide the robot to approach the desired configuration $\bffrakq^*$ at the correct orientation because the nonholonomic constraint prevents making sideways corrections instantaneously. We shape the desired potential energy as:
\begin{equation} \label{eq:desired_potential_energy_1}
\calV_d(\bffrakq_e) = \frac{k_{\bfp}}{2} \bfp_{e}^{\top} \bfp_{e} + \calV_{R_1}(\bffrakq_e)\calV_{R_2}(\bffrakq_e) + \calV_{R_3}(\bffrakq_e),
\vspace*{-0.06cm}
\end{equation}
\begin{equation}\label{eq:desired_potential_energy_terms}
\begin{aligned}
\text{where }\calV_{R_1}(\bffrakq_e) &= \frac{1}{2} \text{tr} \big( \bfI - \bfR_{\Delta_2}^{\top} \bfR(\bfp_e)^{\top} \bfR_{\Delta_1} \bfR^{*} \bfR_e \big), \\ 
\calV_{R_2}(\bffrakq_e) &= \frac{k_{\bfR_1}}{2} \text{tr} \big( \bfI - \bfR^{*\top} \bfR_{\Delta_1}^{\top} \bfR(\bfp_e) \big), \\
\calV_{R_3}(\bffrakq_e) &= \frac{k_{\bfR_2}}{2} \text{tr} \big( \bfI - \bfR(\bfp_e)^{\top} \bfR_{\Delta_1} \bfR^{*} \bfR_{e} \big).
\end{aligned}
\end{equation}
The target-direction rotation matrix, $\bfR(\bfp_e)$, the rotations $\bfR_{\Delta_1}$ and $\bfR_{\Delta_2}$ are defined as:
\begin{gather}
    \bfR(\bfp_e) = 
    \begin{bmatrix}
        \frac{-\bfp_e}{\| \bfp_e \|} & \bfU \frac{-\bfp_e}{\| \bfp_e \|} & \bfe_3
    \end{bmatrix}, \label{eq:Rpe}\\
    \bfR_{\Delta_1} = \bfU \exp \left(\frac{\hat{\bftheta}_1}{\|\bftheta_1\|} \frac{\pi}{2} \right), \;\;
    \bfR_{\Delta_2} = \exp \left(\frac{\hat{\bftheta}_2}{\|\bftheta_2\|} \frac{\pi}{2}\right), \notag 
\end{gather}
where $\bfU = 
        \begin{bmatrix}
            0 & -1 & 0 \\
            1 & 0 & 0 \\ 
            0 & 0 & 1
        \end{bmatrix}$, 
$\bftheta_1 = \log \big( \bfU^{\top} \bfR^{*\top} \bfR(\bfp_e) \big)^{\vee}$, $\bftheta_2 = \log \big( \bfR^{*\top} \bfR_{\Delta_1}^{\top} \bfR(\bfp_e) \big)^{\vee}$, and ${(\cdot)}^{\vee}: \mathfrak{so}(3) \mapsto \mathbb{R}^3$ is the inverse of the hat map. In \eqref{eq:desired_potential_energy_1}, $\calV_d(\bffrakq_e)$ is composed of three components. The first drives the robot position $\bfp$ towards the desired position $\bfp^*$. The second approaches $\bfp^*$ in a circular motion by orienting the robot perpendicular to the direction $-\bfp_e$ and aligning $\bfR(\bfp_e)$ with $\bfR^*$ to specify the rotation direction.
The third aligns $\bfR$ with $\bfR(\bfp_e)$ for a direct straight-line path to $\bfp^*$. Once $\bfR(\bfp_e)$ is in alignment with $\bfR^{*}$, the first component becomes zero, leaving only the second component active. 
\begin{figure*}[t]
    \captionsetup{justification=centering}
    \subcaptionbox{Loss (log scale)\label{fig:loss_sim}}{\includegraphics[width=0.23\linewidth]{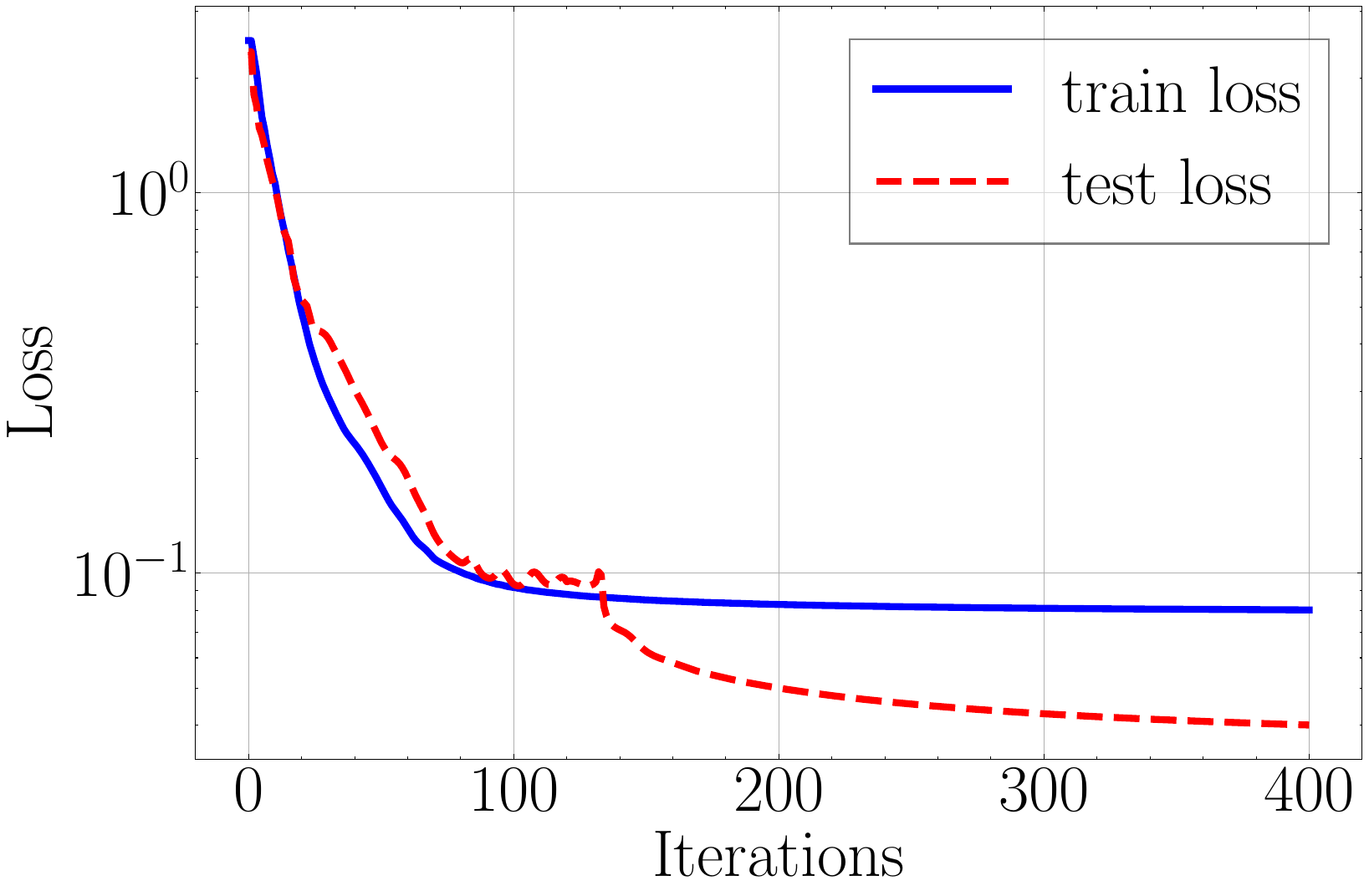}}%
    \hfill%
    \subcaptionbox{$SO(3)$ constraints\label{fig:SO3_constraints_sim}}{\includegraphics[width=0.23\linewidth]{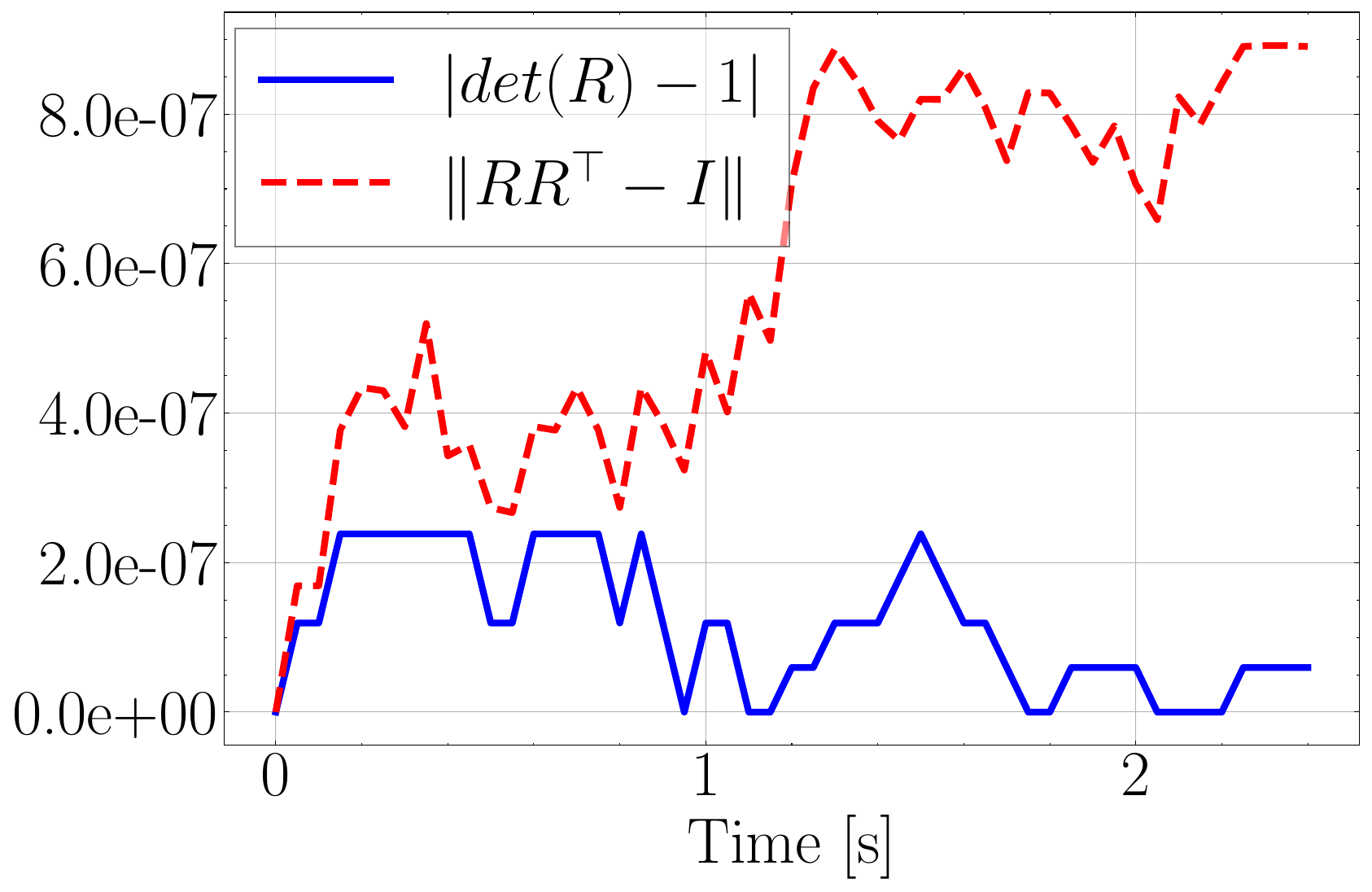}}%
    \hfill%
    \subcaptionbox{Inversed mass $\bfM_{\bftheta}^{-1}$\label{fig:M_sim}}{\includegraphics[width=0.22\linewidth]{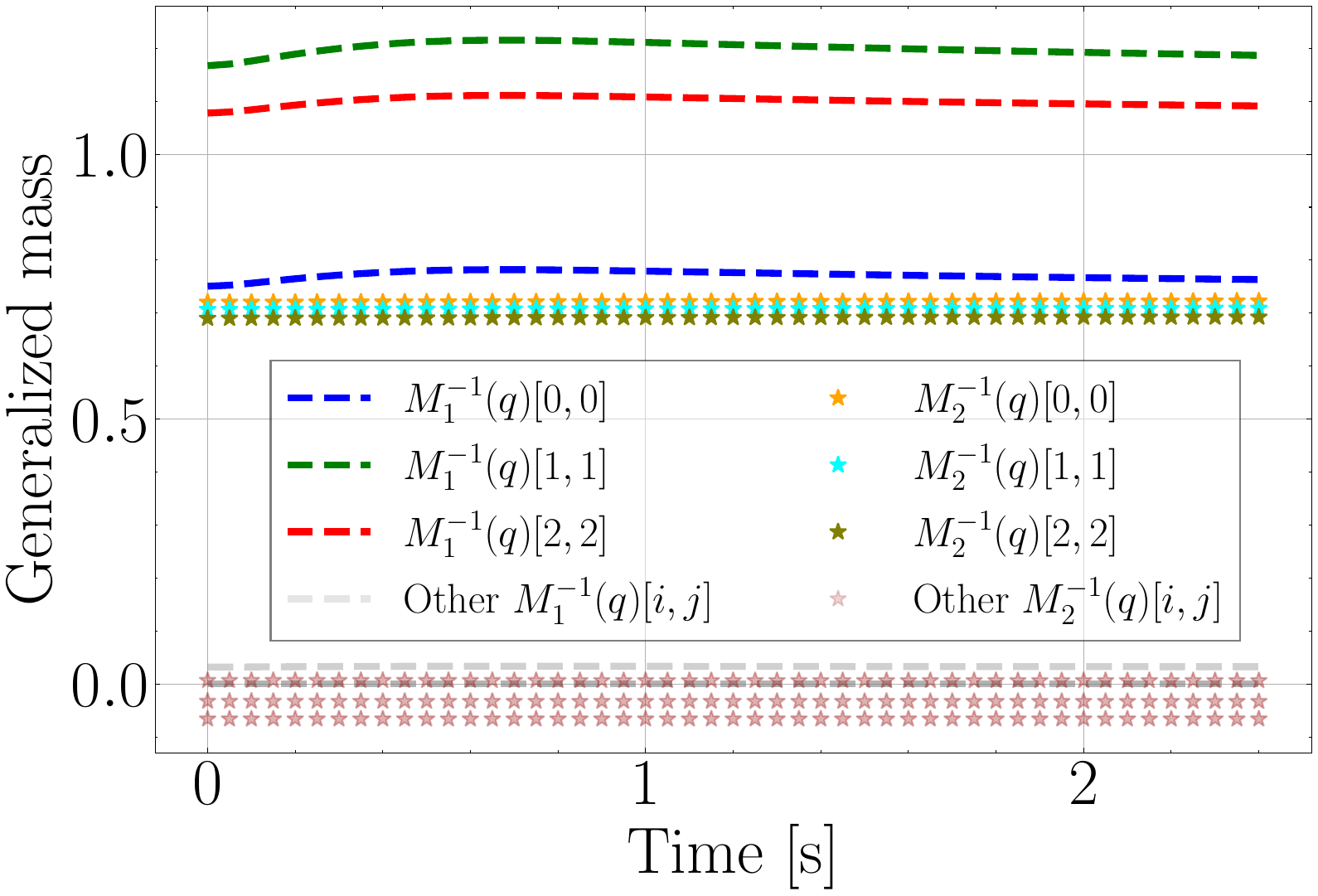}}%
    \hfill%
    \subcaptionbox{Input gain $\bfg_{\bftheta}$\label{fig:g_sim}}{\includegraphics[width=0.23\linewidth]{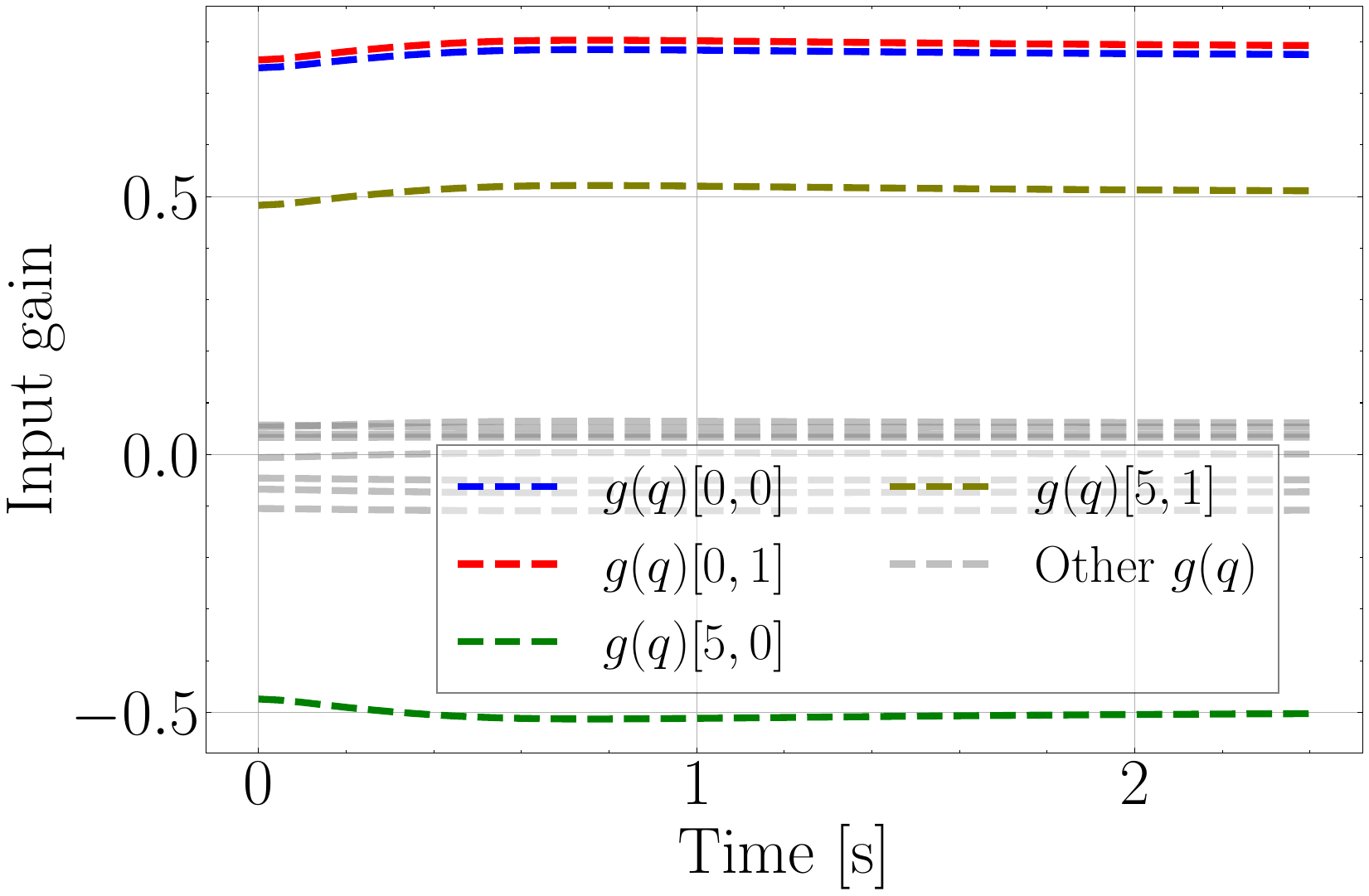}}\\
    \subcaptionbox{Dissipation matrix $\bfD_{\bfv,\bftheta}$\label{fig:Dv_sim}}{\includegraphics[width=0.22\linewidth]{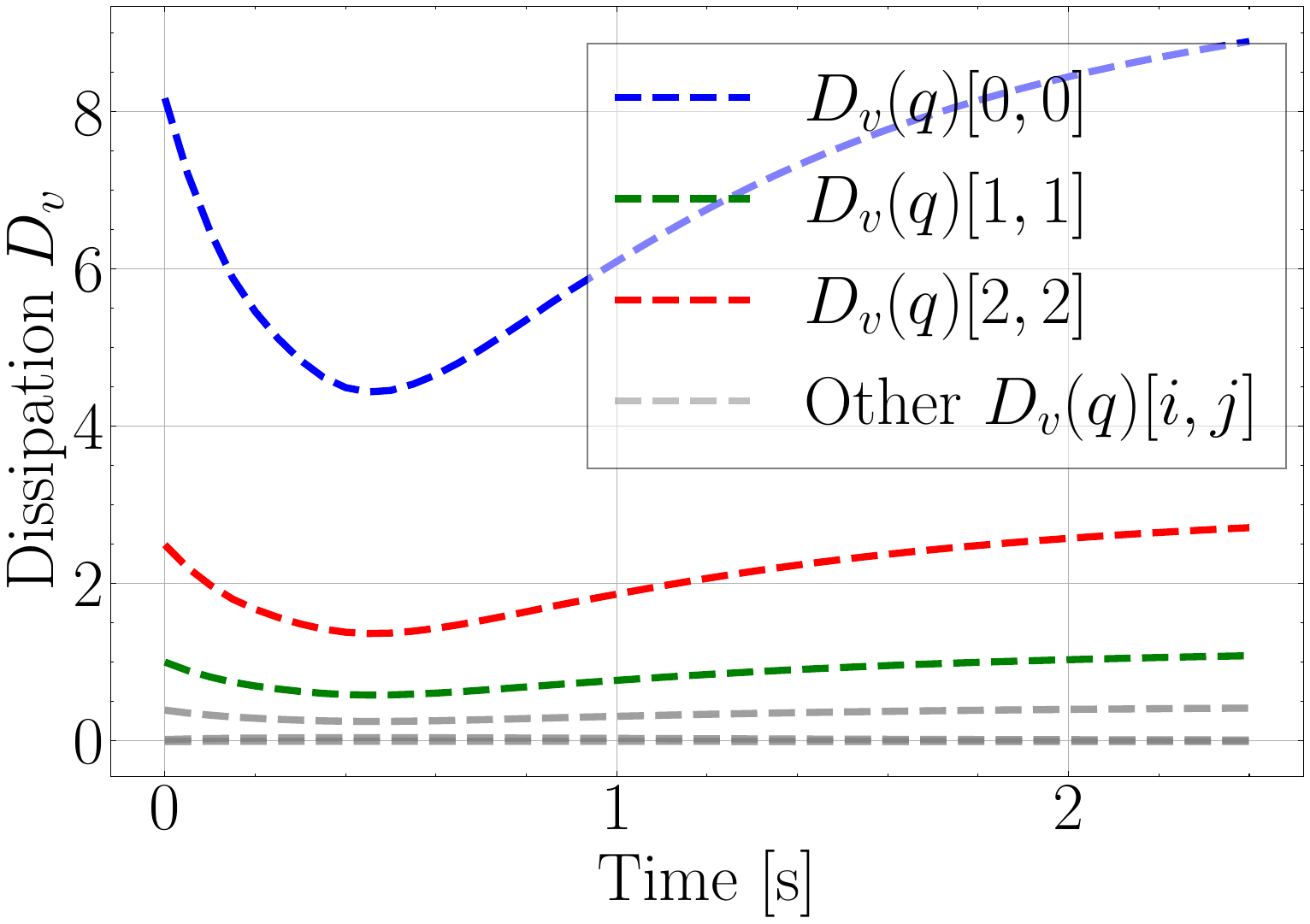}}%
    \hfill%
    \subcaptionbox{{Dissipation matrix $\bfD_{\bfomega, \bftheta}$}\label{fig:Dw_sim}}{\includegraphics[width=0.225\linewidth]{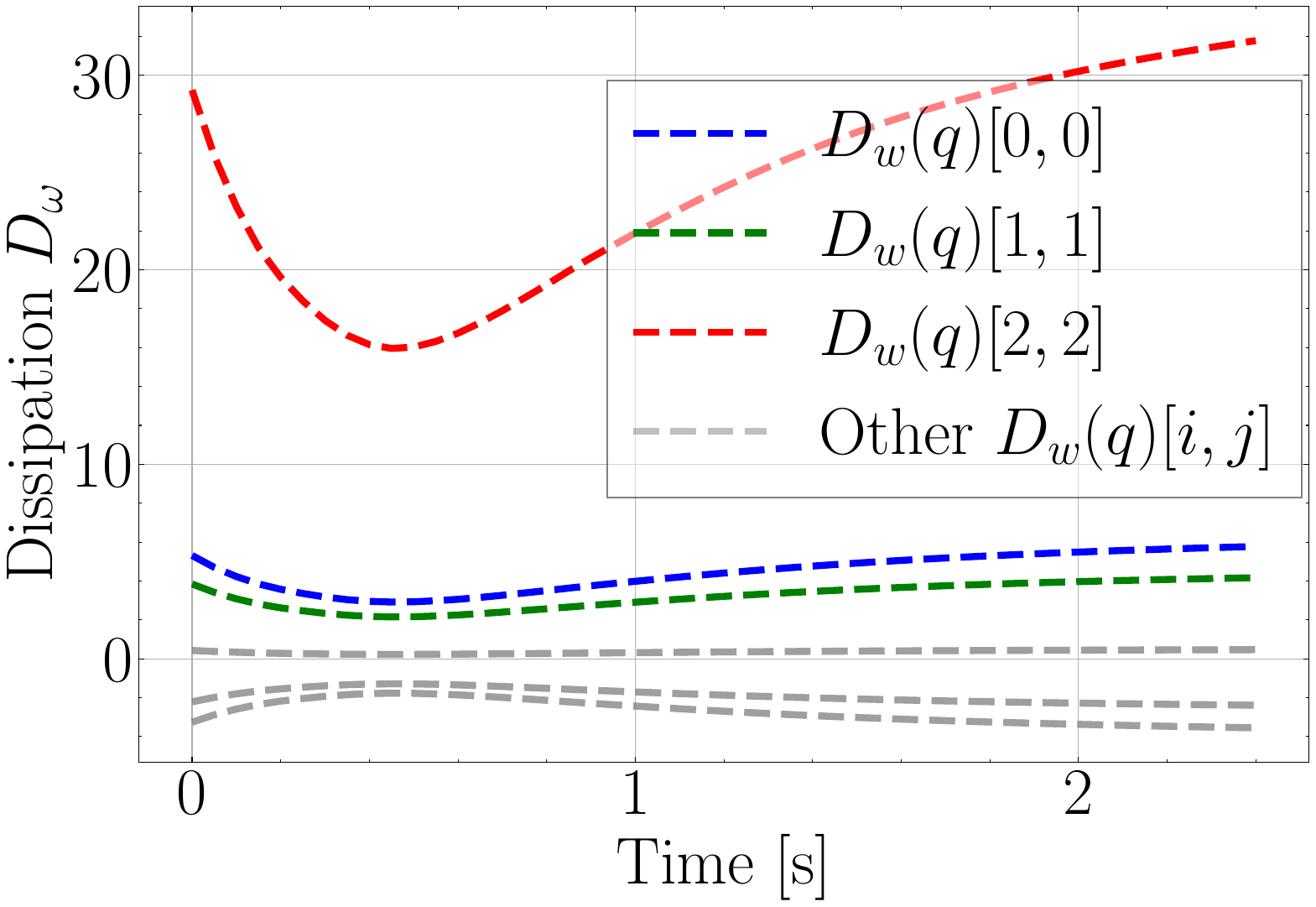}}%
    \hfill%
    \subcaptionbox{Potential energy $\calV_{\bftheta}$\label{fig:V_sim}}{\includegraphics[width=0.23\linewidth]{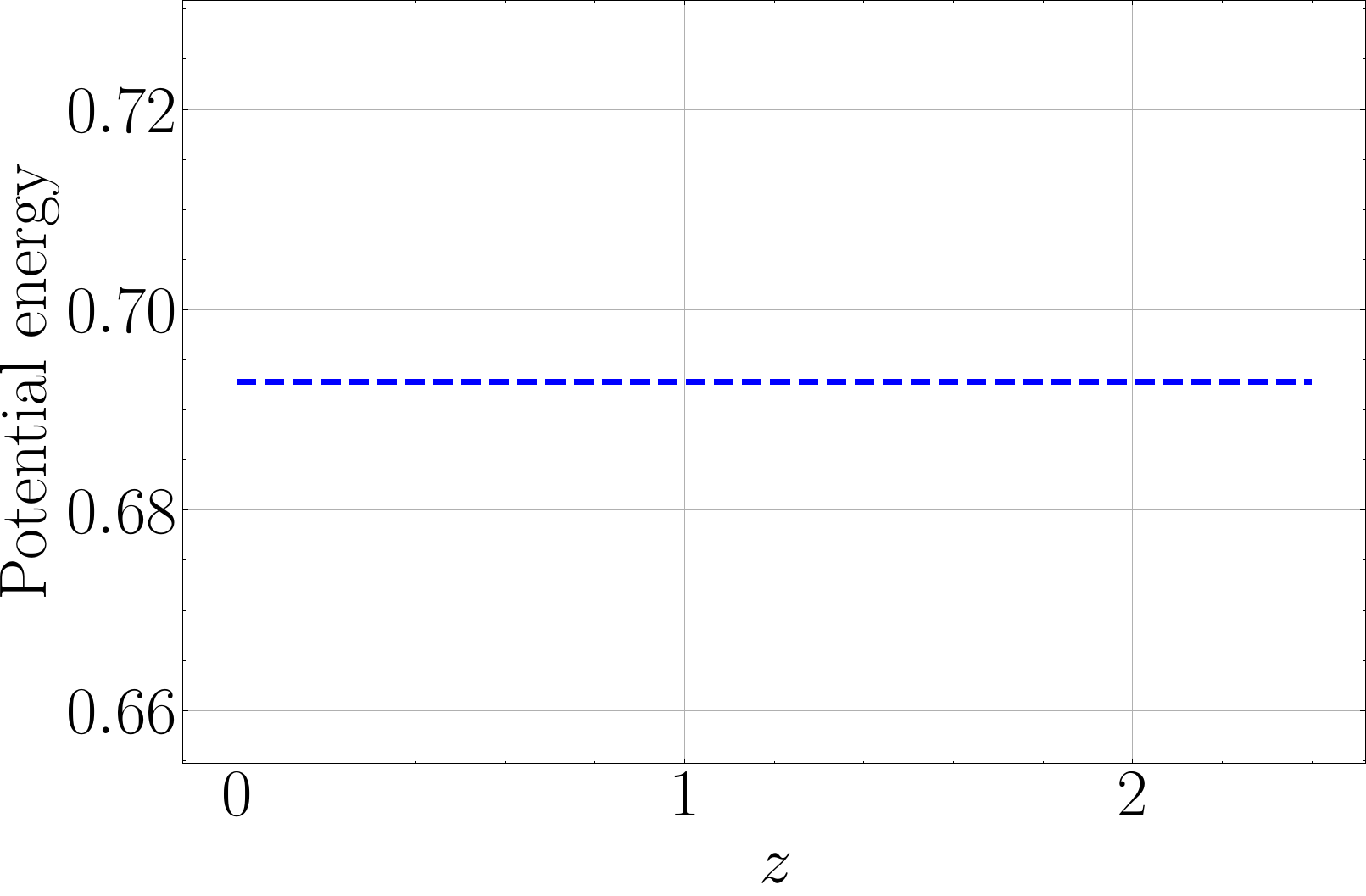}}%
    \hfill%
    \subcaptionbox{Pose stabilization in simulation\label{fig:sim_control_plots}}{\includegraphics[width=0.24\linewidth]{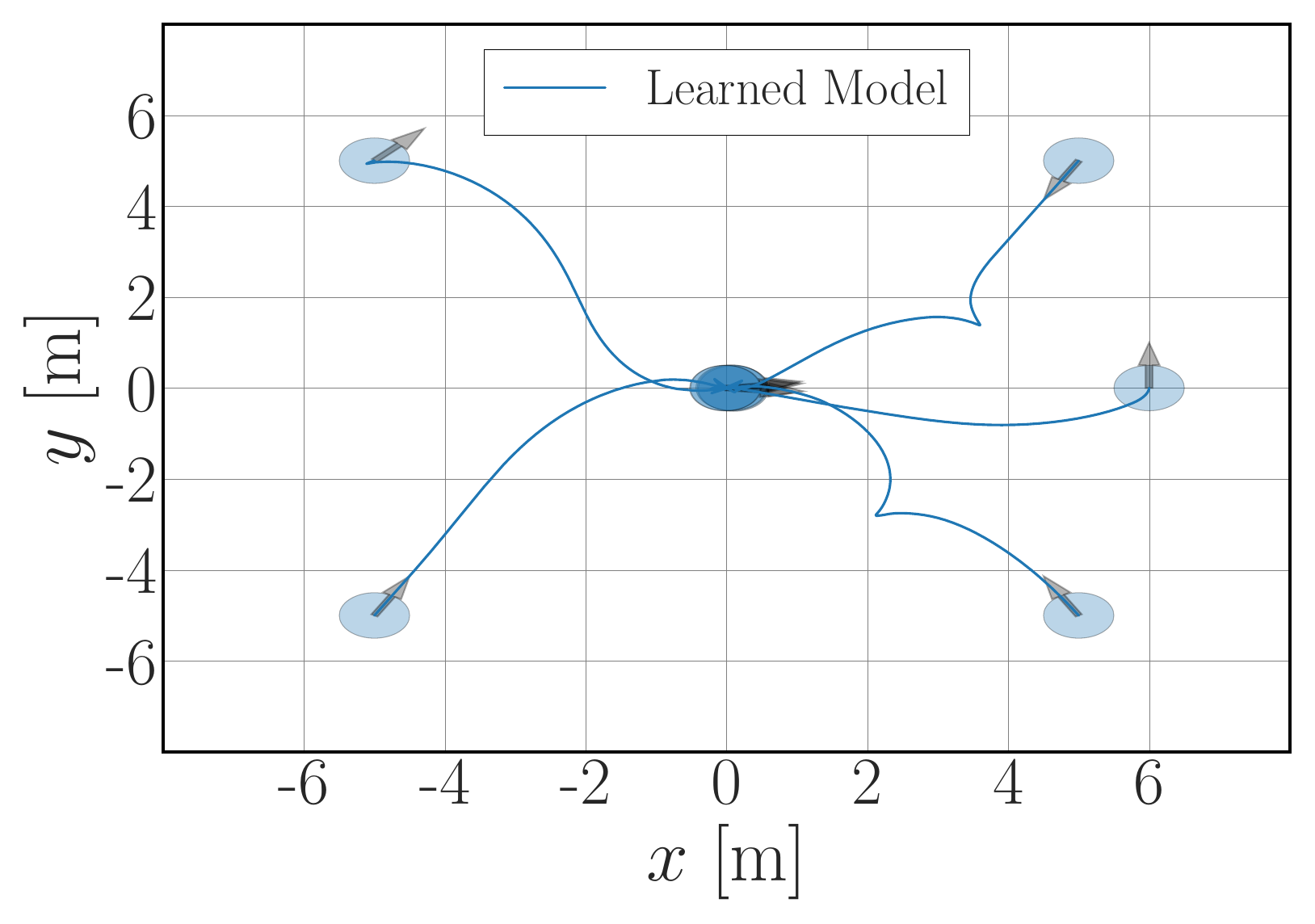}}
    \caption{Evaluation of our Hamiltonian Neural ODE network (a)-(g) along a trajectory and (h) pose stabilization in the Sapien simulator.}
    \label{fig:simulation_results}
    \vspace*{-0.35cm}
\end{figure*}

Assuming a constant $\bfM_{\bftheta}(\bffrakq) \equiv \bfM_{\bftheta}$, Eq. \eqref{eq:u_ES_DI} becomes: 
\begin{align}
    \bfu_{ES} &= \bfg (\boldsymbol{\bffrakq})^{\dagger} \bigg( \boldsymbol{\bffrakq}^{\times \top} \frac{\partial \calV_{\bftheta}}{\partial \boldsymbol{\bffrakq}} - \boldsymbol{\bffrakp}^{\times}\bfM_{\bftheta}^{-1} \boldsymbol{\bffrakp} + \bfD_{\bftheta}(\boldsymbol{\bffrakq}, \boldsymbol{\bffrakp}) \bfM_{\bftheta}^{-1} \boldsymbol{\bffrakp} \notag \\ 
    & \qquad \qquad  -\bfe(\boldsymbol{\bffrakq}, \boldsymbol{\bffrakq}^{*}) + \dot{\bfp}^{*} \bigg), \label{eq:u_es_and_u_di}\\ 
    \bfu_{DI} &= -\bfg (\boldsymbol{\bffrakq})^{\dagger} \bfK_{\bfd} \bfM_{\bftheta}^{-1} (\boldsymbol{\bffrakp} - \boldsymbol{\bffrakp}^{*}), \notag
\end{align}
with $\bfe(\boldsymbol{\bffrakq}, \boldsymbol{\bffrakq}^{*}) = \begin{bmatrix} \bfe_{\bfv}(\boldsymbol{\bffrakq}, \boldsymbol{\bffrakq}^{*})^{\top} & \bfe_{\bfomega}(\boldsymbol{\bffrakq}, \boldsymbol{\bffrakq}^{*})^{\top} \end{bmatrix}^{\top}$ defined as:
\begin{align}\label{eq:e_v_and_e_w}
    &\bfe_{\bfv}(\boldsymbol{\bffrakq}, \boldsymbol{\bffrakq}^{*}) = \bfR^{\top} \bigg( k_{\bfp} \bfp_e + \calV_{R_2}(\boldsymbol{\bffrakq}_e) \frac{\partial \calV_{R_1}(\boldsymbol{\bffrakq}_e)}{\partial \bfp_e} \notag\\
    & \qquad\qquad\qquad + \calV_{R_1}(\boldsymbol{\bffrakq}_e) \frac{\partial \calV_{R_2}(\boldsymbol{\bffrakq}_e)}{\partial \bfp_e} + \frac{\partial \calV_{R_3}(\boldsymbol{\bffrakq}_e)}{\partial \bfp_e} \bigg),\\ 
    &\bfe_{\bfomega}(\boldsymbol{\bffrakq}, \boldsymbol{\bffrakq}^{*}) = \frac{k_{\bfR_2}}{2} \bigg( \bfR(\bfp_e)^{\top} \bfR_{\Delta_1} \bfR - \bfR^{\top} \bfR_{\Delta_1}^{\top} \bfR(\bfp_e)  \bigg)^{\vee} \notag\\
    &+\frac{\calV_{R_2}(\boldsymbol{\bffrakq}_e)}{2} \bigg(\! \bfR_{\Delta_2}^{\top} \bfR(\bfp_e)^{\top} \bfR_{\Delta_1} \bfR - \bfR^{\top} \bfR_{\Delta_1}^{\top} \bfR(\bfp_e) \bfR_{\Delta_2} \!\bigg)^{\vee}\!\!.\notag
\end{align}
Please refer to the extended version\footnote{Paper website: \url{https://altwaitan.github.io/DLFO/}.} of the paper \cite{dlfo_extended} for the computation of the derivatives of the terms $\calV_{R_i}$.

\section{Evaluation}
\label{sec:experiment}

In this section, we verify the effectiveness of our approach in both simulated and real experiments with a Jackal robot. We describe data collection for training in Sec. \ref{subsec:data_collection}, and then evaluate our training results and control performance in simulation (Sec. \ref{subsec:simulation}) and in real experiments (Sec. \ref{subsec:real_experiment}).

\subsection{Data Collection}
\label{subsec:data_collection}
We collected a dataset $\calD = \{t_n^{(i)}, \calY_n^{(i)}, \bfx_0^{(i)}, \bfu_n^{(i)}\}_{i=1,n=0}^{D,N}$ consisting of $D$ $N$-sample trajectories with sampling interval of $dt$ seconds by driving the robot manually using a joystick. The convergence of model parameter estimates requires a persistence of excitation condition to be satisfied \cite{persistenceofexcitation}. Hence, it is important to collect a diverse range of linear and rotational motions.
A short trajectory duration $N$ is chosen to avoid numerical instability of the predicted states from an ODE solver. Each data point includes a point cloud $\calY_{n}^{(i)}$ from a 3D LiDAR sensor and the control inputs $\bfu_{n}^{(i)}$. From two consecutive LiDAR scans, we apply fast global registration~\cite{zhou2016fast} to pick $M$ random correspondences to create two corresponding point clouds $\calY_n^{(i)}$ and $\calY_{n+1}^{(i)}$. 

\begin{figure*}[t]
    \captionsetup{justification=centering}
    \subcaptionbox{Loss (log scale)\label{fig:loss_sim_real}}{\includegraphics[width=0.23\linewidth]{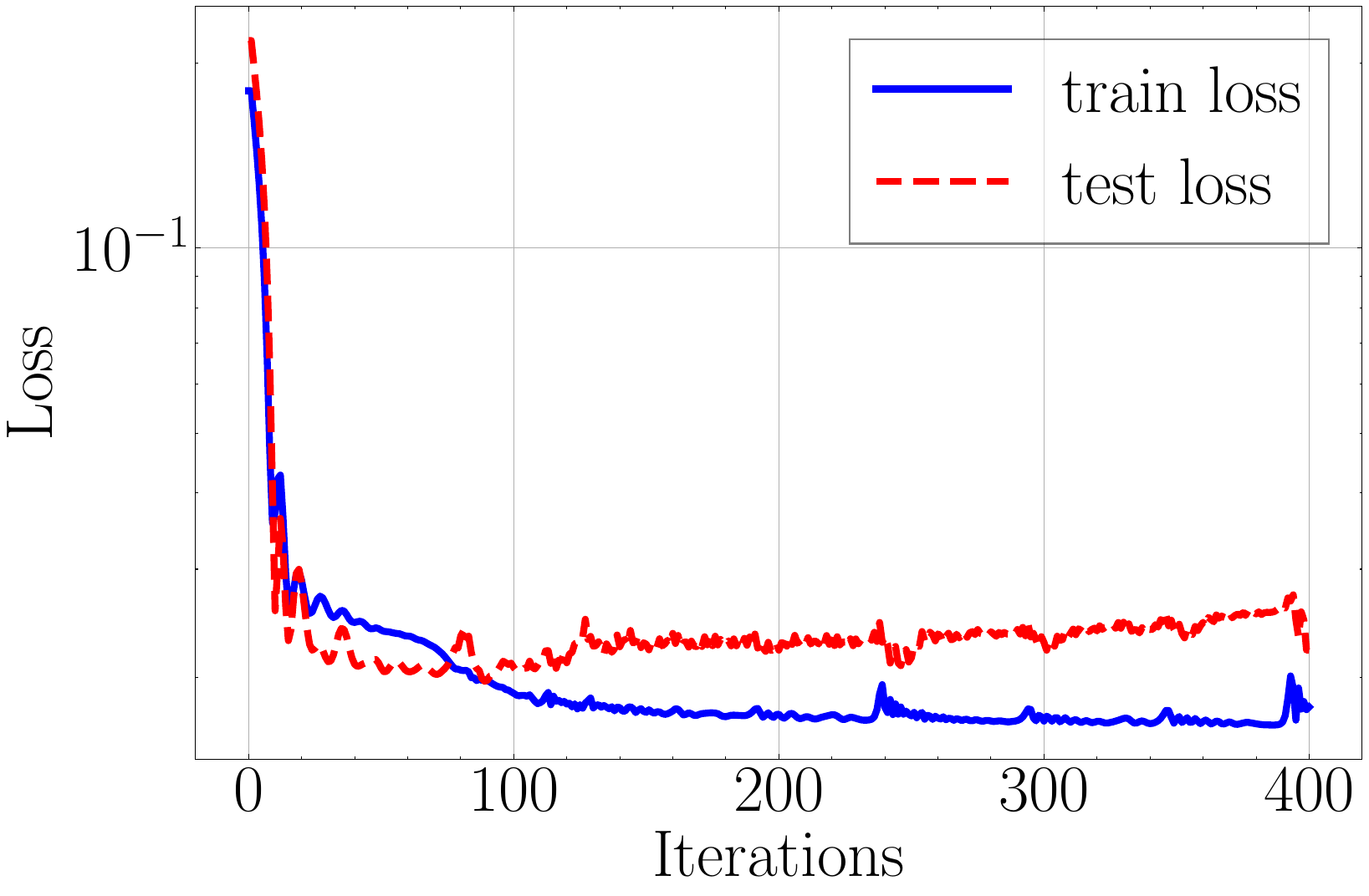}}%
    \hfill%
    \subcaptionbox{$SO(3)$ constraints\label{fig:SO3_constraints_sim_real}}{\includegraphics[width=0.23\linewidth]{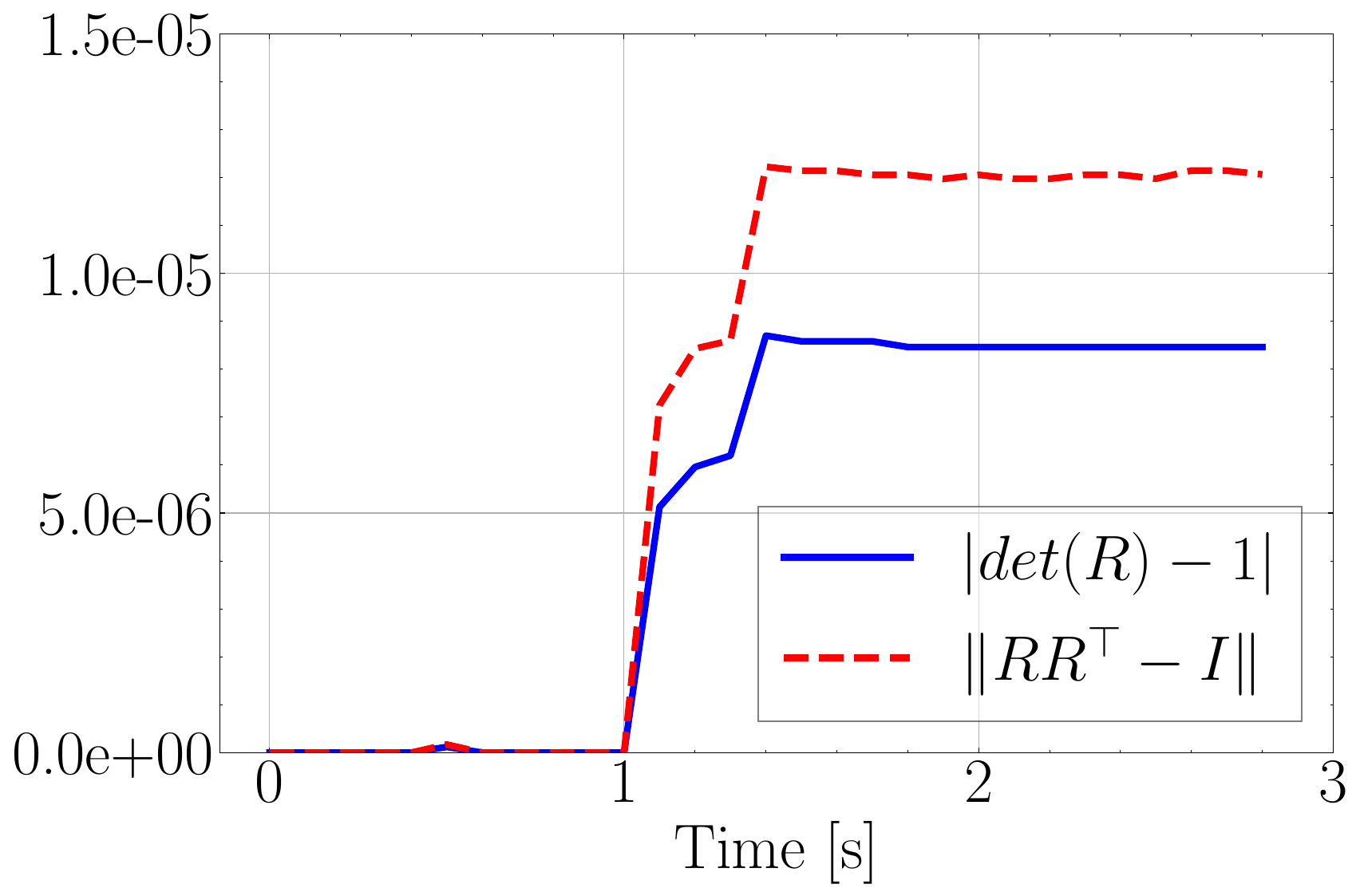}}%
    \hfill%
    \subcaptionbox{Inverse mass $\bfM_{1,\bftheta}^{-1}$\label{fig:M1_sim_real}}{\includegraphics[width=0.22\linewidth]{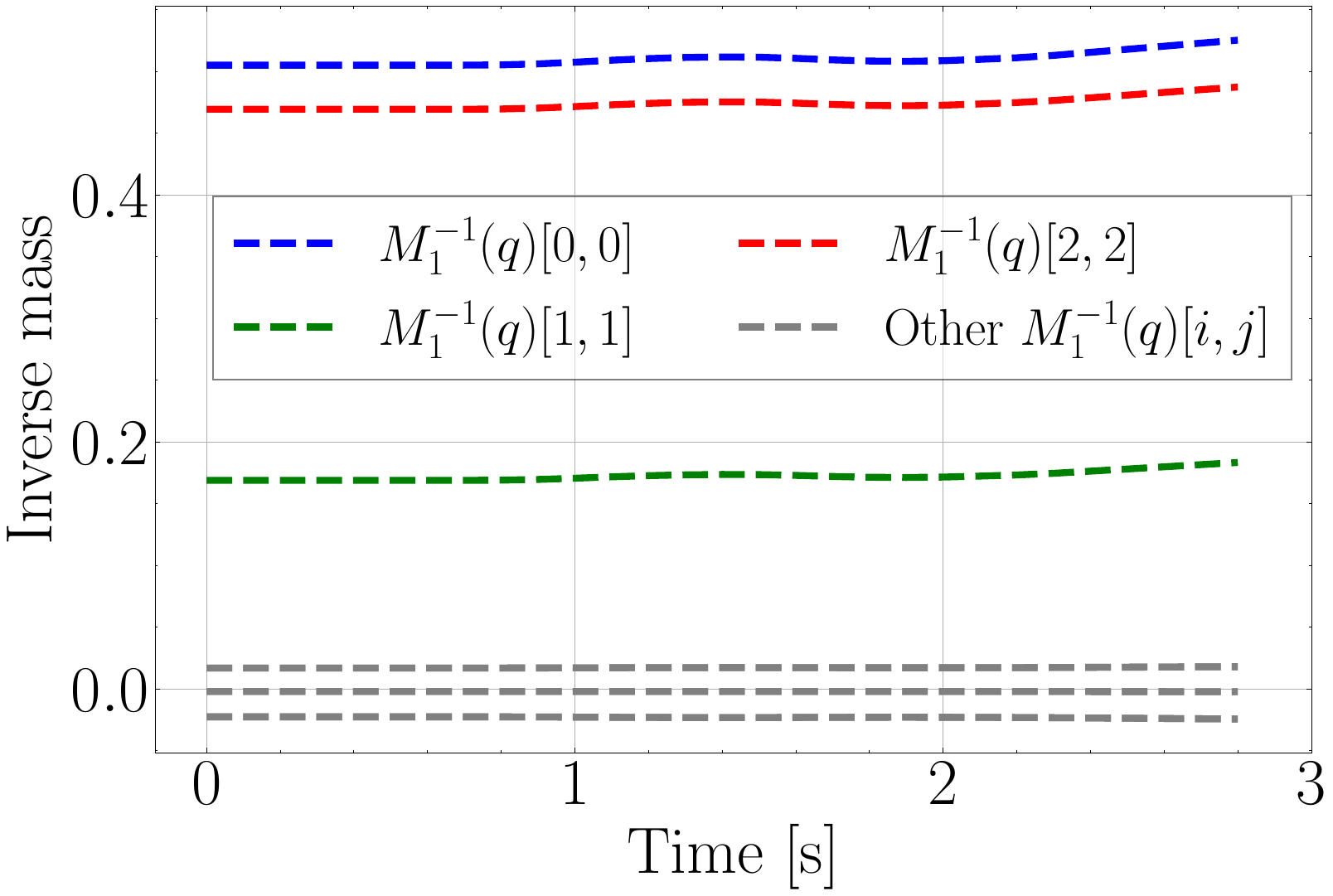}}%
    \hfill%
    \subcaptionbox{Inverse inertia $\bfM_{2,\bftheta}^{-1}$\label{fig:M2_sim_real}}{\includegraphics[width=0.22\linewidth]{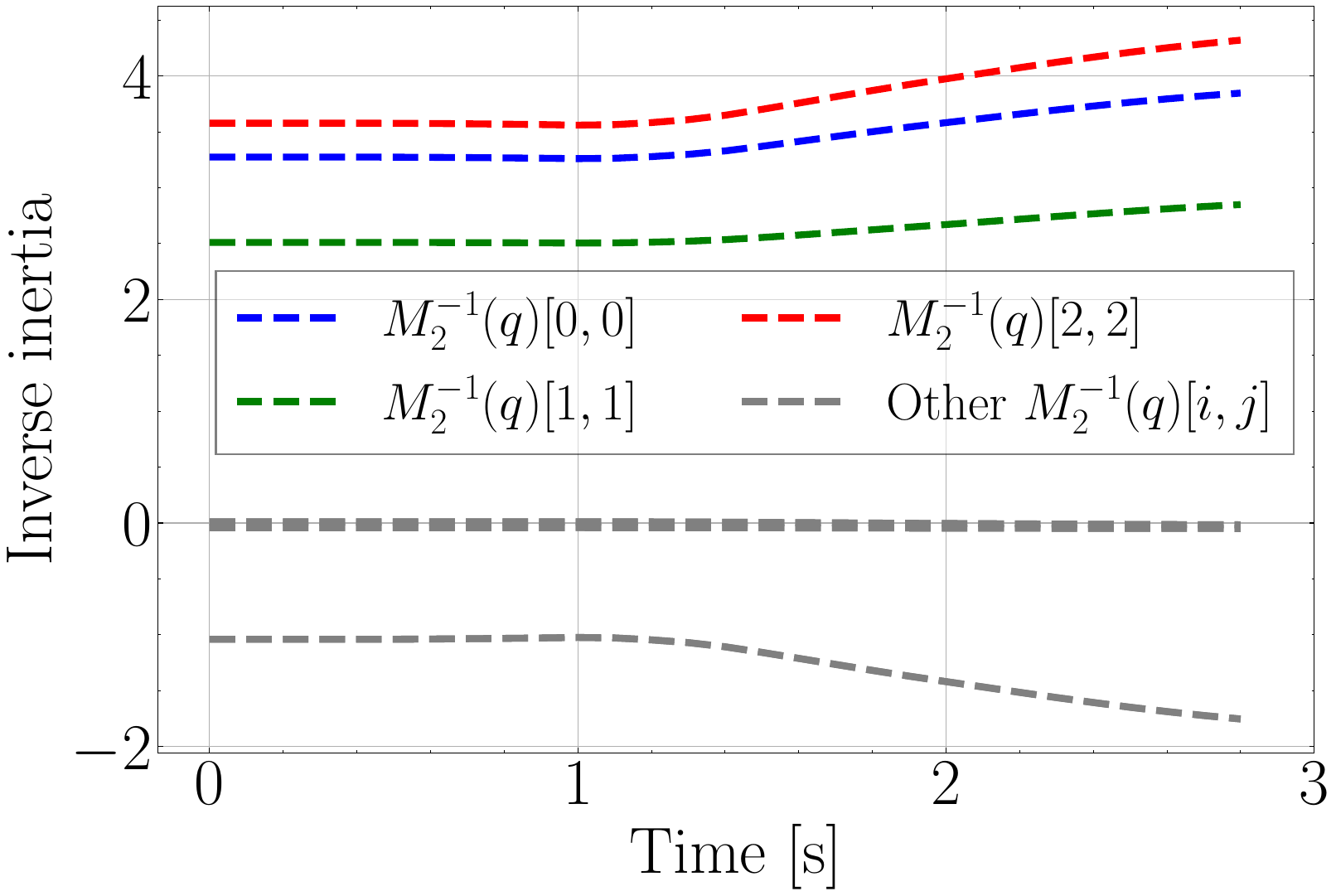}}%
    \hfill%
    \subcaptionbox{Input gain $\bfg_{\bftheta}$\label{fig:g_sim_real}}{\includegraphics[width=0.22\linewidth]{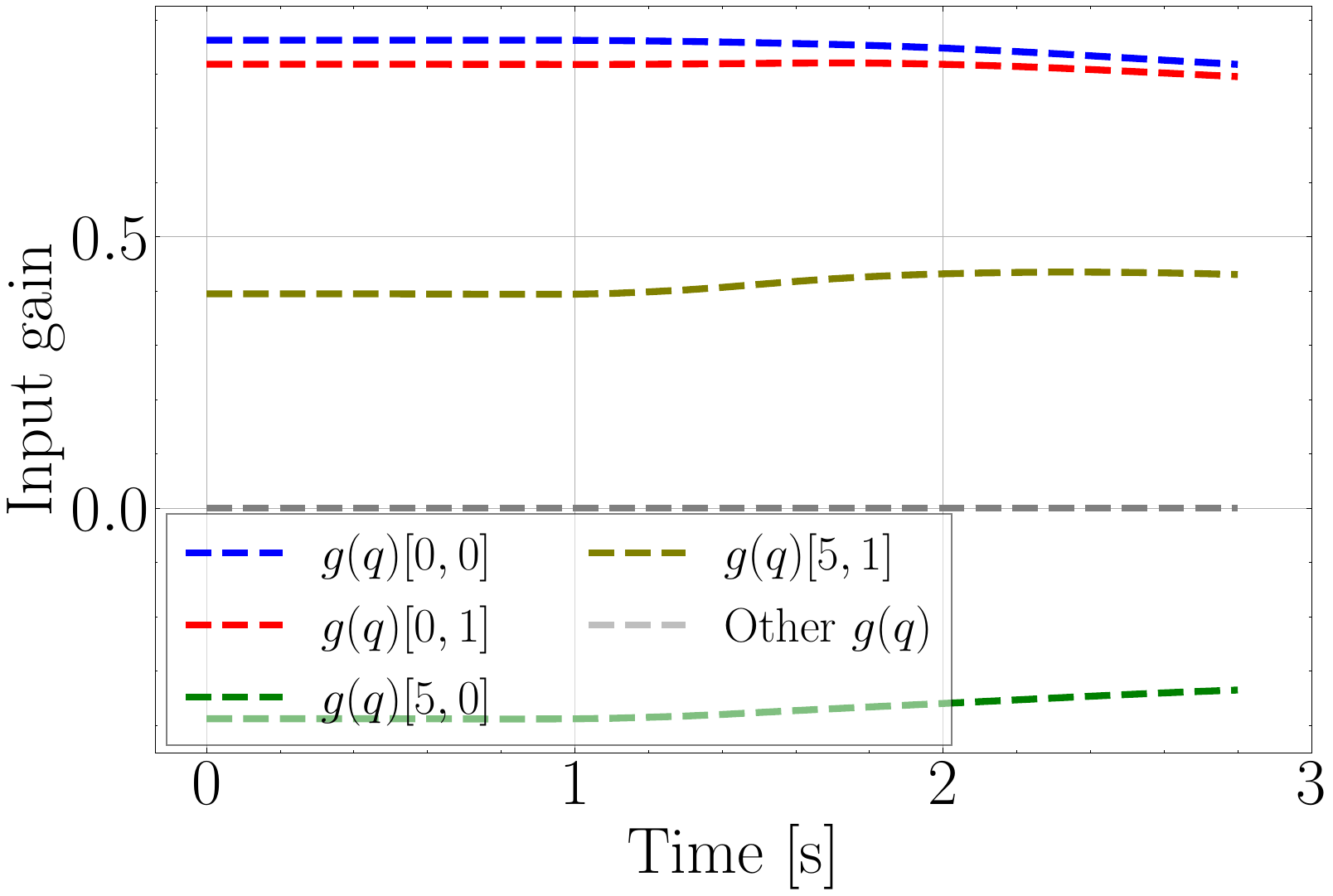}}%
    \hfill%
    \subcaptionbox{Dissipation matrix $\bfD_{\bfv,\bftheta}$\label{fig:Dv_sim_real}}{\includegraphics[width=0.22\linewidth]{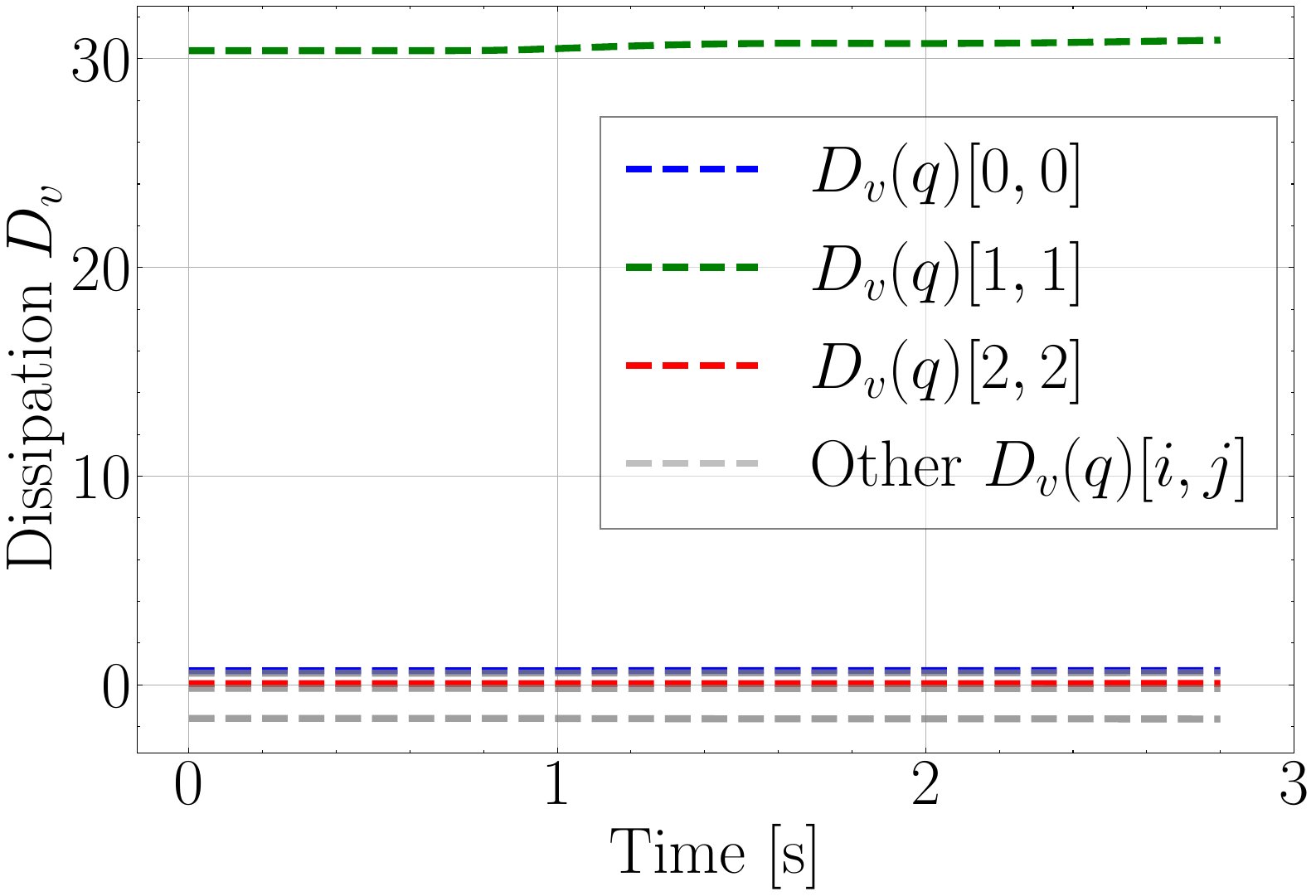}}%
    \hfill%
    \subcaptionbox{Dissipation matrix $\bfD_{\bfomega, \bftheta}$\label{fig:Dw_sim_real}}{\includegraphics[width=0.22\linewidth]{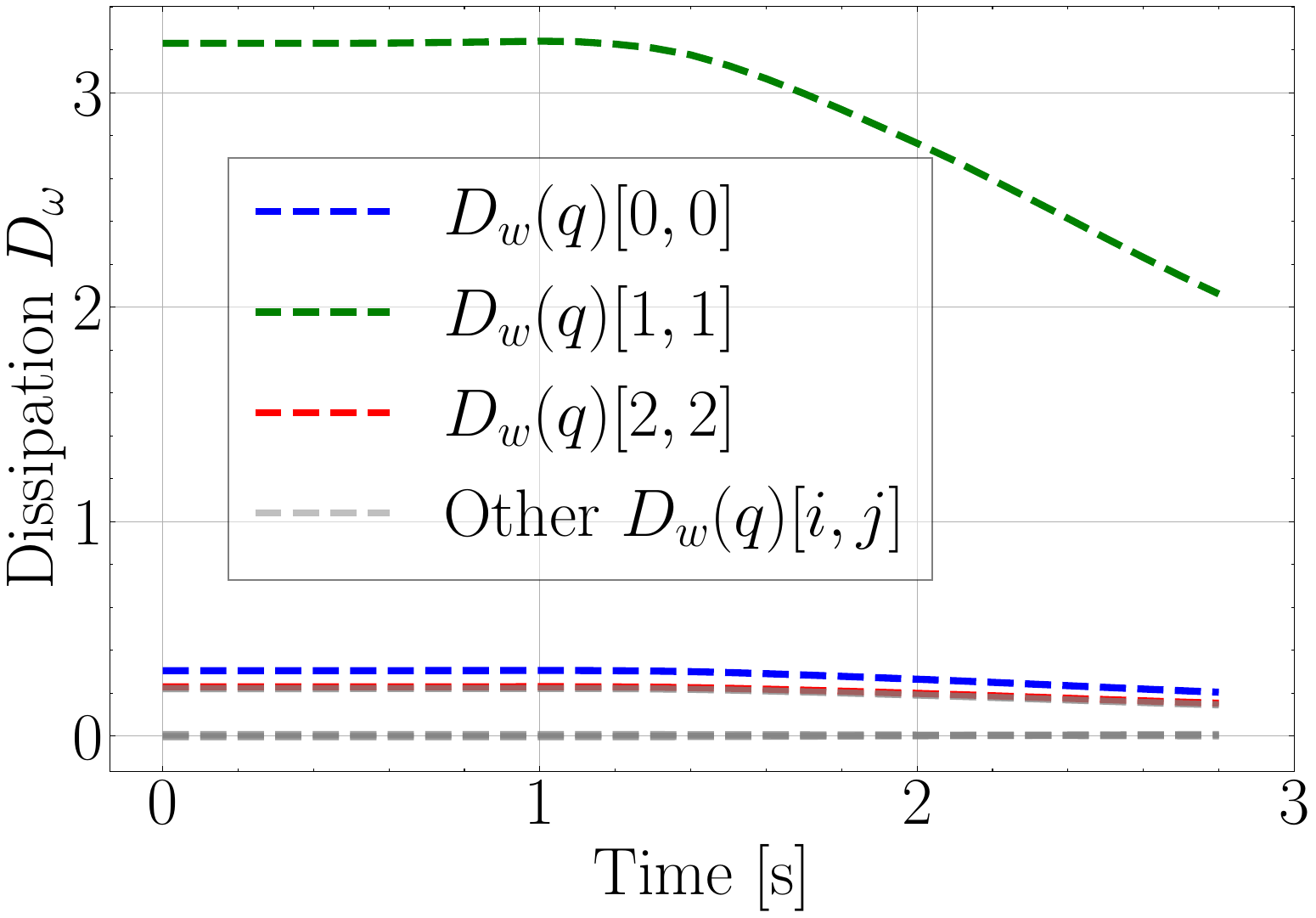}}%
    \hfill%
    \subcaptionbox{Pose stabilization on real Jackal\label{fig:sim_control_plots_real}}{\includegraphics[width=0.235\linewidth]{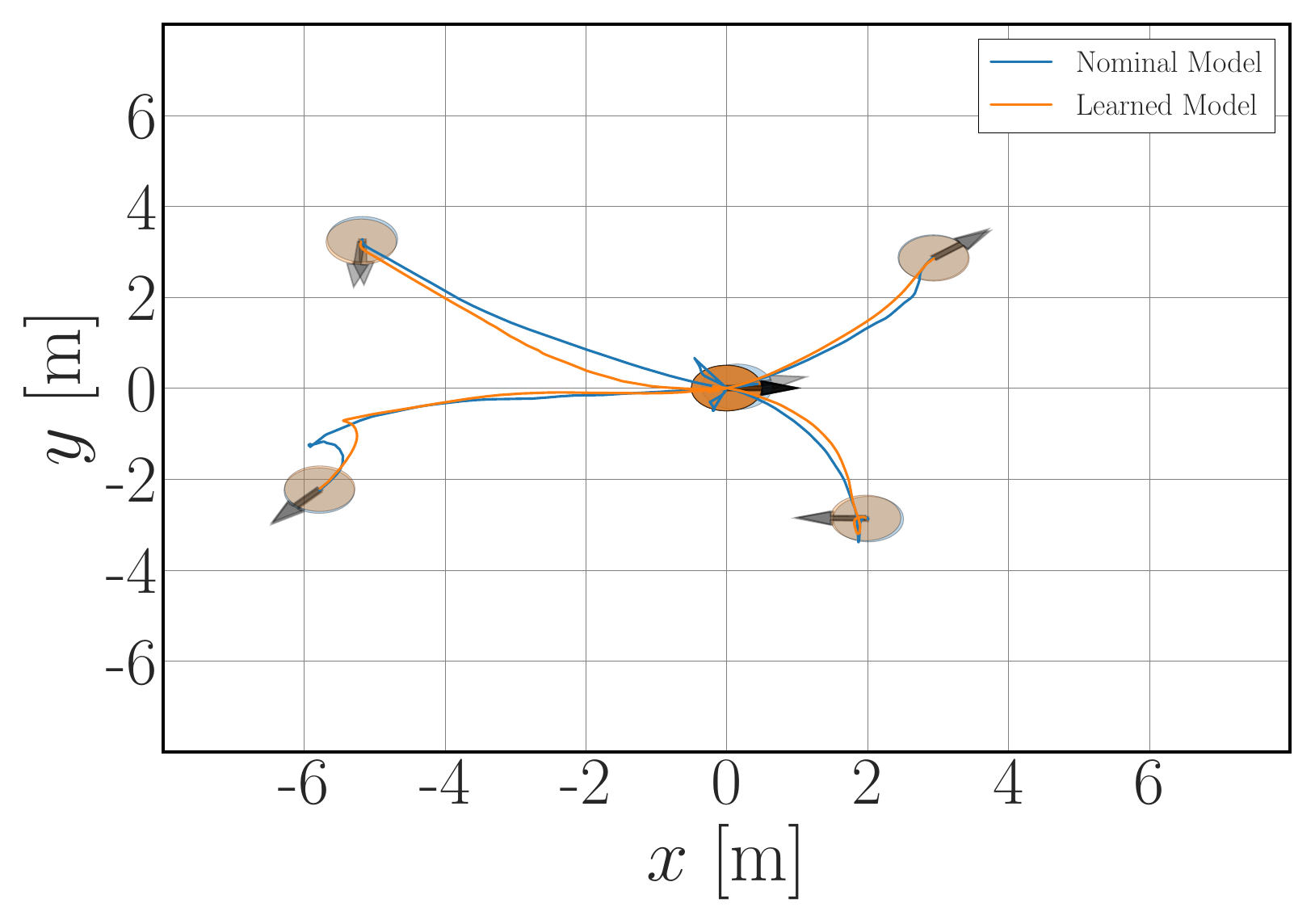}}
    \caption{Evaluation of our Hamiltonian neural ODE network (a)-(g) along a trajectory and (h) pose stabilization on a real Jackal.}
    \label{fig:real_jackal_learned_model}
    \vspace*{-0.25cm}
\end{figure*}
\begin{figure*}[t]
    \captionsetup{justification=centering}
    \subcaptionbox{Lemniscate (nominal) \label{fig:nom_lemniscate}}{\includegraphics[width=0.22\linewidth]{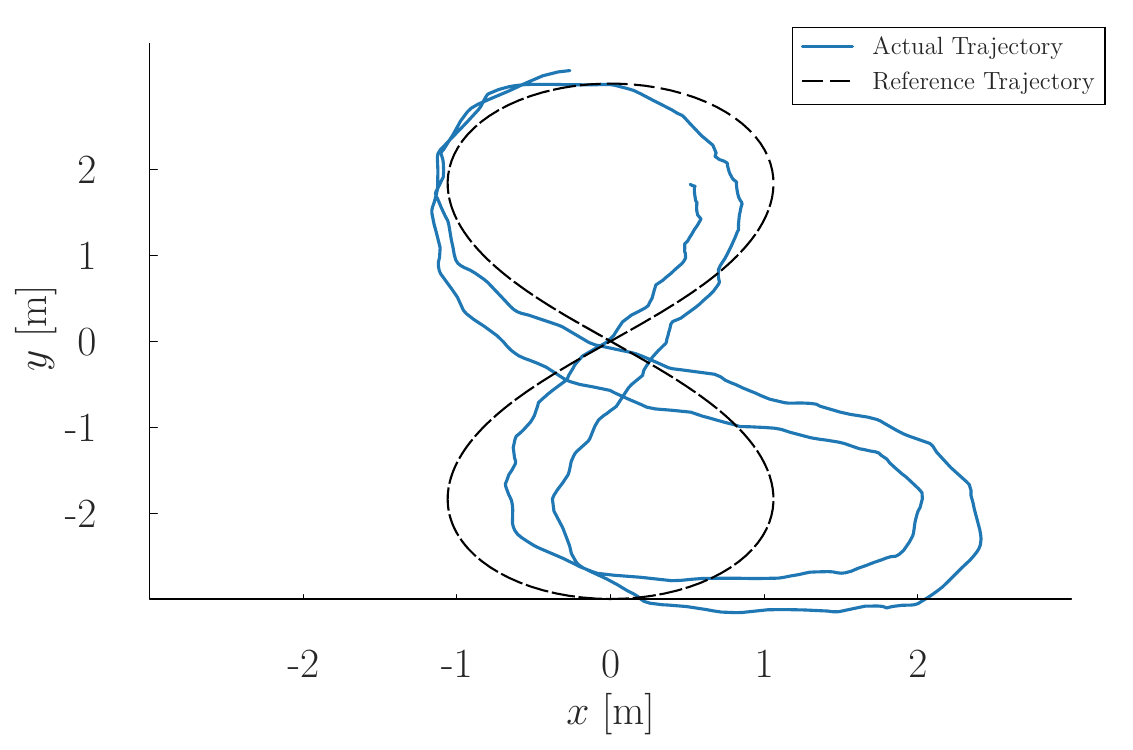}}%
    \hfill%
    \subcaptionbox{Circle (nominal) \label{fig:nom_circle}}{\includegraphics[width=0.22\linewidth]{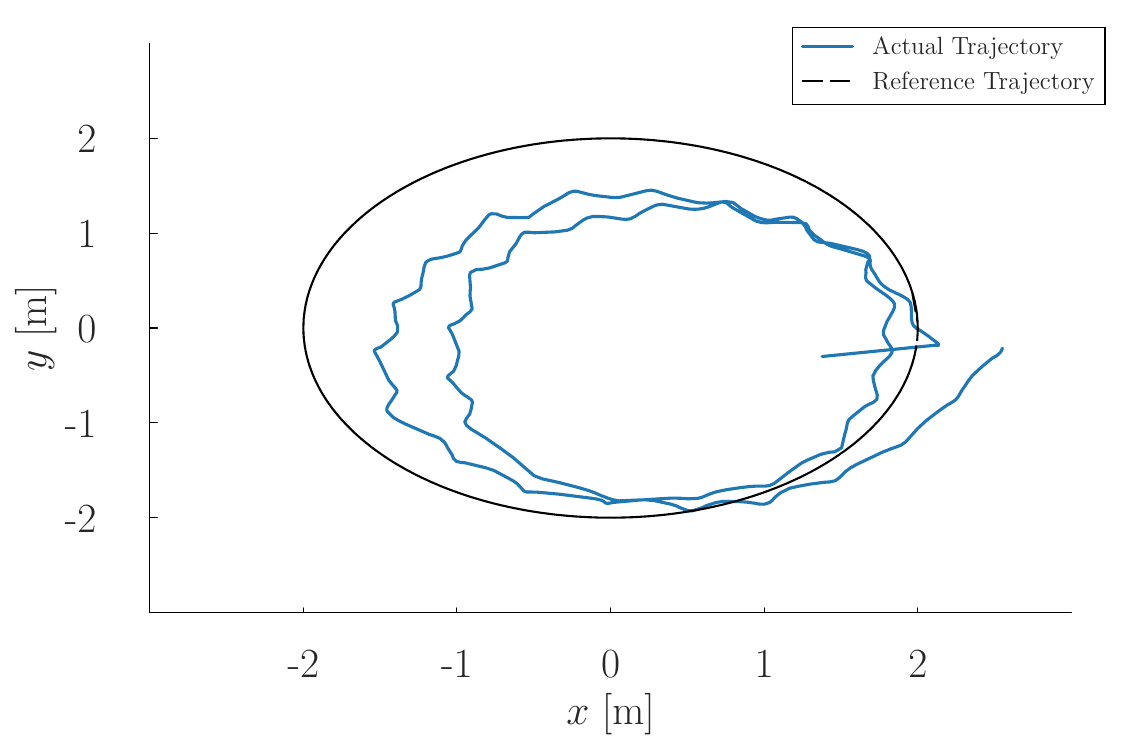}}%
    \hfill%
    \subcaptionbox{$x, y, \theta$ (circle, nominal) \label{fig:nom_circle_xyth}}{\includegraphics[width=0.23\linewidth]{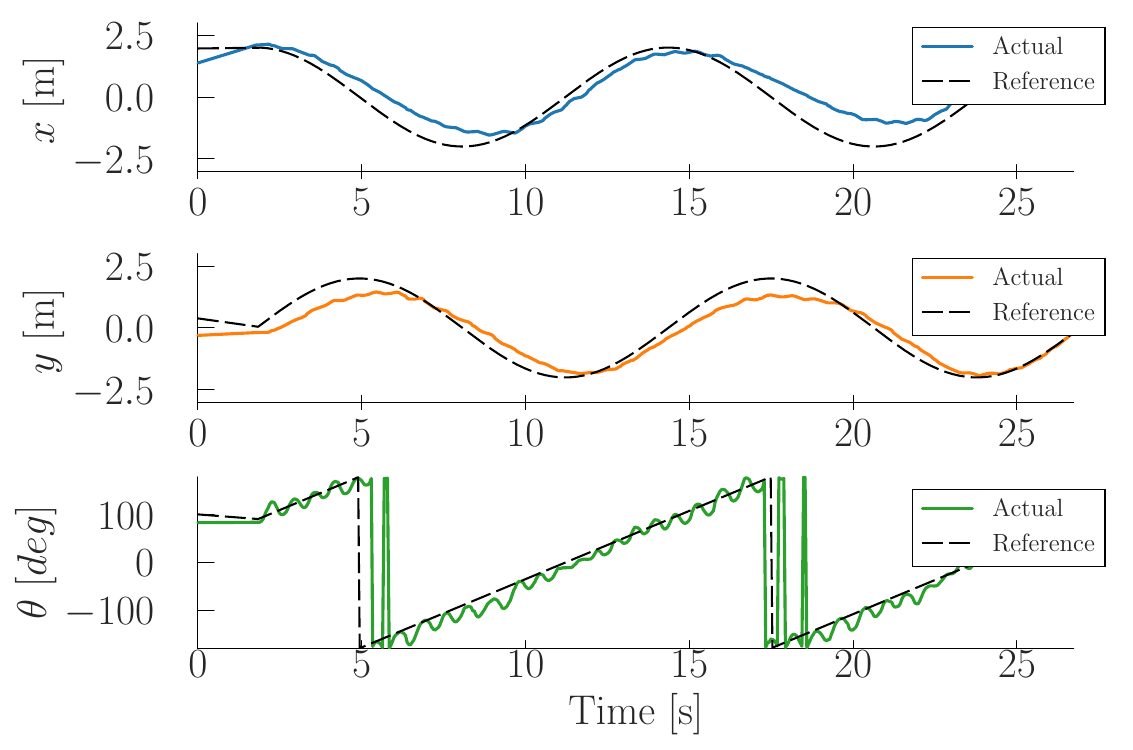}}%
    \hfill%
    \subcaptionbox{Velocities (circle, nominal)  \label{fig:nom_circle_vel}}{\includegraphics[width=0.23\linewidth]{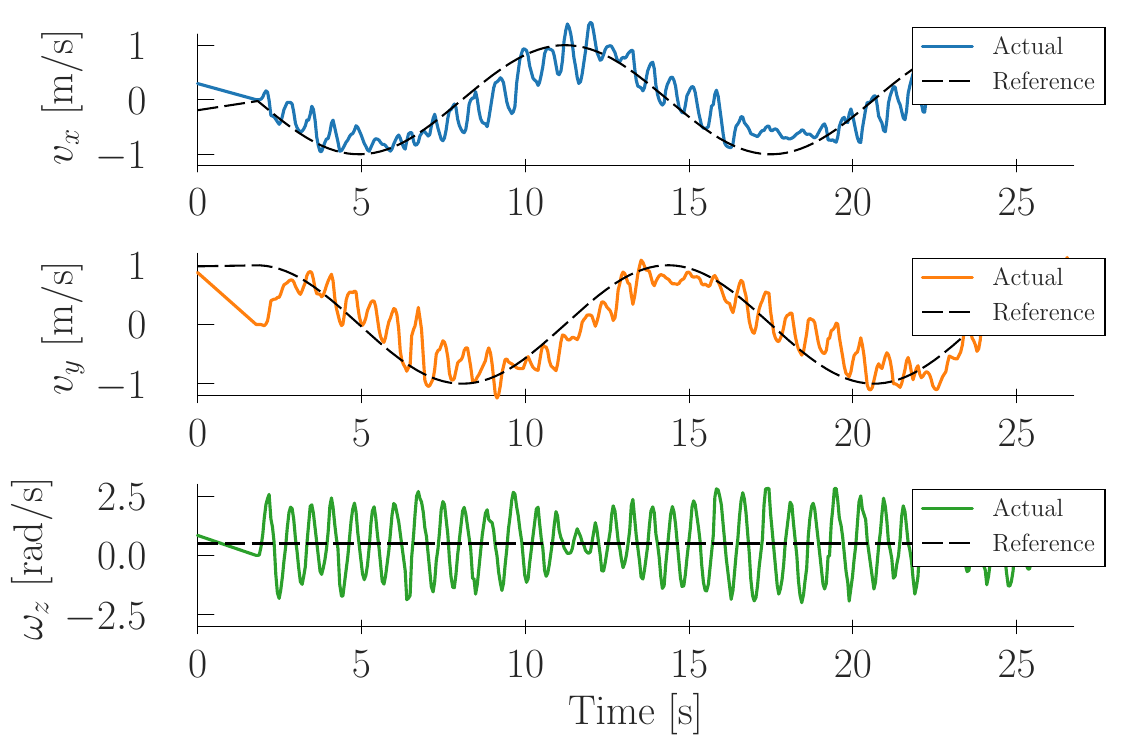}}\\
    \subcaptionbox{Lemniscate (learned)  \label{fig:learned_lemniscate}}{\includegraphics[width=0.22\linewidth]{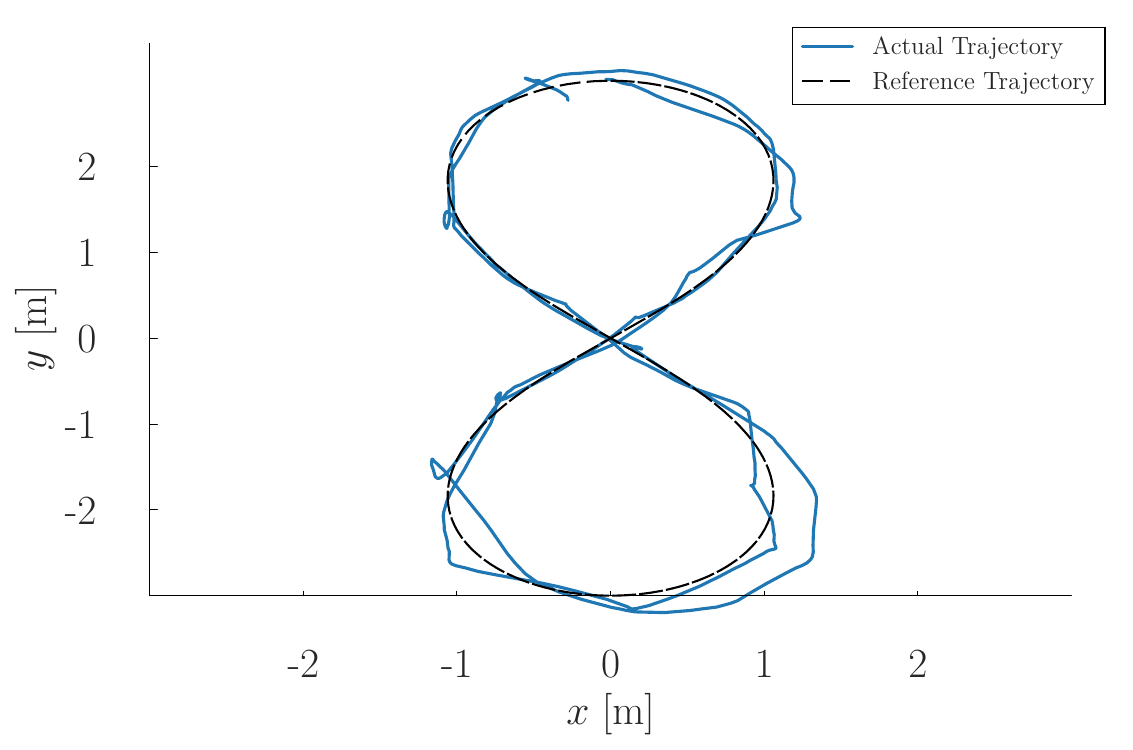}}%
    \hfill%
    \subcaptionbox{Circle (learned)  \label{fig:learned_circle}}{\includegraphics[width=0.22\linewidth]{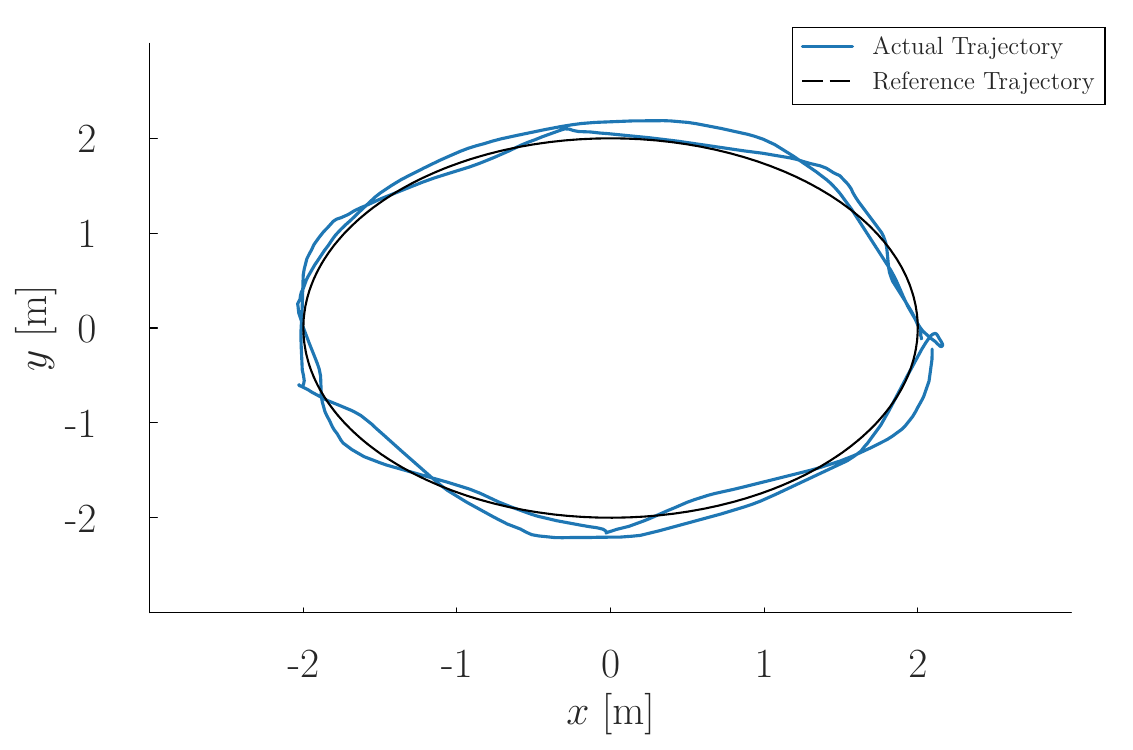}}%
    \hfill%
    \subcaptionbox{$x, y, \theta$ (circle, learned) \label{fig:learned_circle_xyth}}{\includegraphics[width=0.23\linewidth]{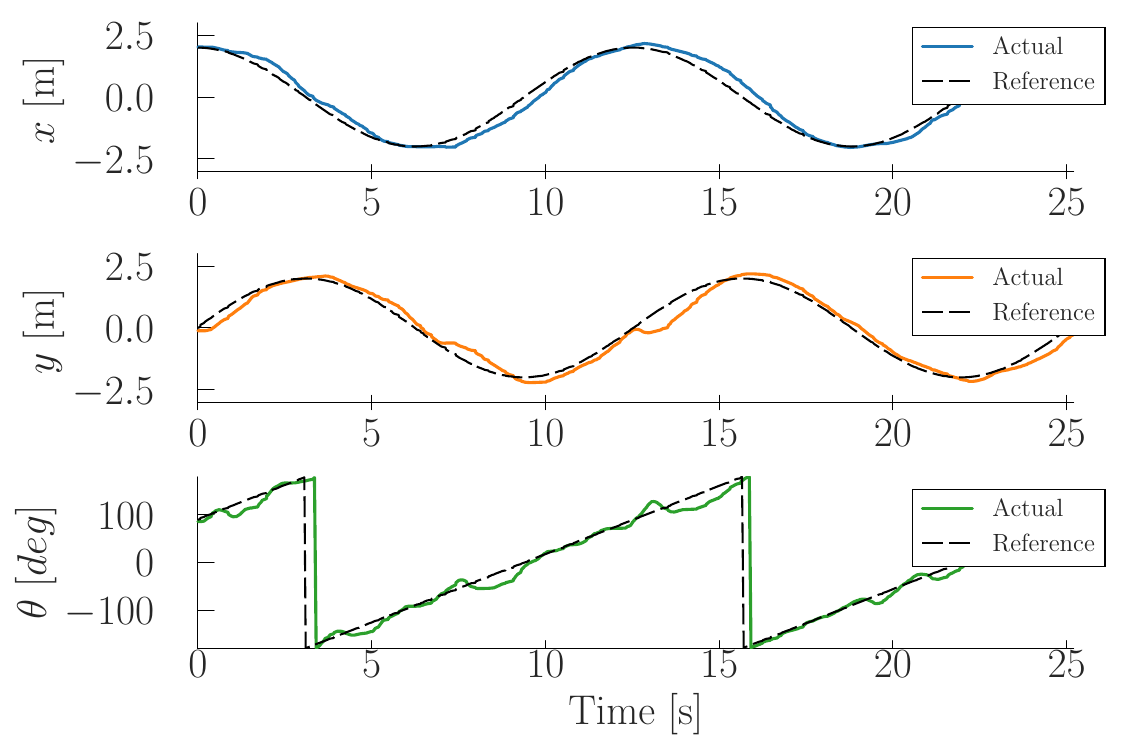}}%
    \hfill%
    \subcaptionbox{Velocities  (circle, learned)\label{fig:learned_circle_vel}}{\includegraphics[width=0.23\linewidth]{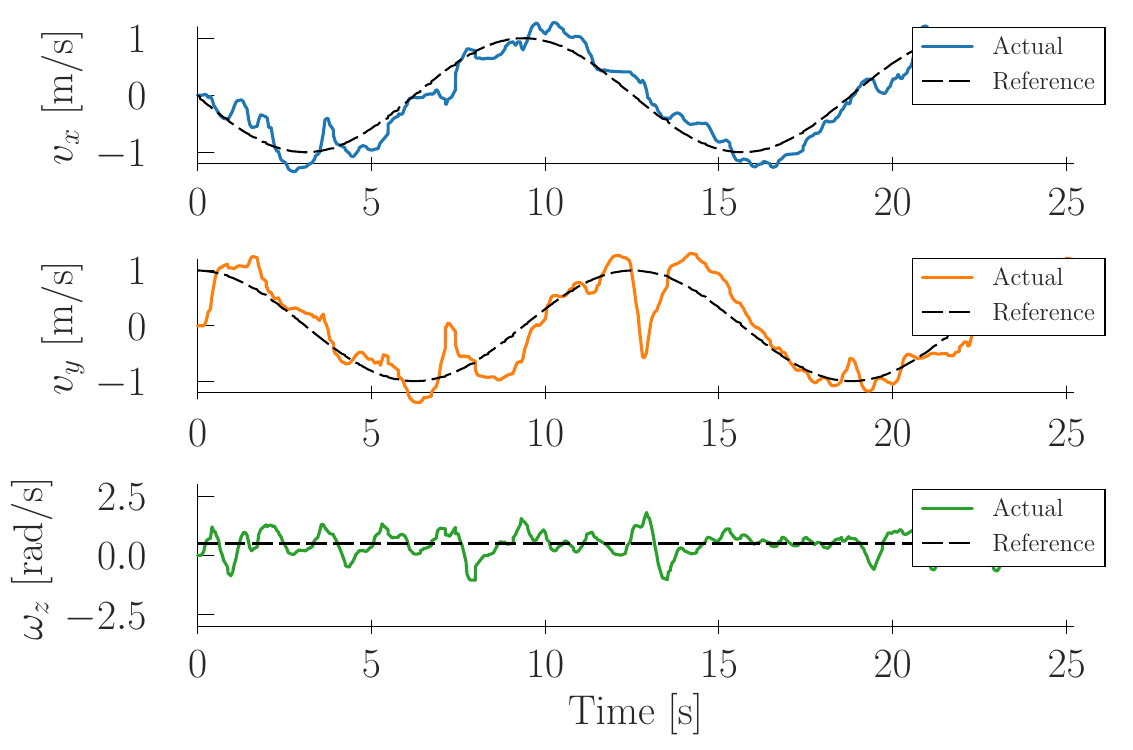}}
    \caption{Evaluating nominal and learned model performance in tracking circular and lemniscate trajectories.}
    \label{fig:tracking}
    \vspace*{-0.25cm}
\end{figure*}
%
\subsection{Sapien Simulation}
\label{subsec:simulation}
We build a simulated differential drive robot in a physics-based simulation environment Sapien \cite{Xiang_2020_SAPIEN}. The control input is $\bfu^{(i)}_n = \begin{bmatrix} \tau_L & \tau_R \end{bmatrix}^{\top}  \in \mathbb{R}^2$, consisting of left $\tau_L$ and right $\tau_R$ wheel torques. A dataset $\calD$ was collected as described in Sec. \ref{subsec:data_collection} with $(N, D, M, dt) = (250, 32, 40, 0.05)$ by driving the robot using sampled wheel torques.

We train our dynamics model as described in Sec. \ref{sec:learning_using_cycle_consistency} with the dataset $\calD$ for $400$ iterations. The networks are pretrained to a nominal model $\bfM_0(\bffrakq) = \bfI$, $\bfD_0(\bffrakq) = 10^{-3}\bfI$, $\calV_0(\bffrakq) = 0$, $\bfg_0(\bffrakq) = \begin{bmatrix}
1 & 0 & 0 & 0 & 0 & -1 \\ 
1 & 0 & 0 & 0 & 0 & 1
\end{bmatrix}^{\top}$ to avoid numerical instability by the ODE solver. The regularization loss $\ell_{r}({\bftheta})$ is set to $10^{-3}||\bfg_{\bftheta}(\bffrakq)||_1$ to encourage sparsity in $\bfg_{\bftheta}(\bffrakq)$. Fig. \ref{fig:simulation_results} shows training and testing performance for our HNODE design. The $SO(3)$ constraints are satisfied, by design, and the potential energy $\calV_{\bftheta}(\bffrakq)$ remains constant due to 2D plane motion. The learned mass $\bfM_{\bftheta}(\bffrakq)$ converged to a diagonal matrix and the input gain $\bfg_{\bftheta}(\bffrakq)$ is sparse with nearly equal values in its first row and opposite signs in the last, consistent with the nonholonomic constraint.

The controller proposed in Section \ref{subsec:ida_pbc_differential_drive} was designed with the learned parameters and control gain values $(k_{\bfp}, k_{\bfR_1}, k_{\bfR_2}, \bfK_{\bfd}) = (1.2, 7, 3, \text{diag}(1.2 \bfI, \ \bfI))$. Fig. \ref{fig:sim_control_plots} shows that our controller successfully drives the robot to the desired pose from various initial poses with desired yaw achieved before converging to desired position.

\subsection{Real World Deployment}
\label{subsec:real_experiment}
We use a Clearpath Jackal robot to demonstrate our approach. The control input $\bfu = \begin{bmatrix} \Delta \omega_1 & \Delta \omega_2  \end{bmatrix}^{\top} \in \mathbb{R}^2$ represents the difference between the desired and actual wheel velocity $\Delta \omega_i = \omega^{*}_{i} - \omega_{i}$. The closed-loop dynamics of the wheels are modeled as a fist order system, similar to the approach in \cite{Faessler17},  with $\dot{\omega} = \frac{1}{\alpha} (\omega^{*} - \omega)$, where $\alpha$ denotes the system's time constant. In doing so, we establish a means to control the wheel's angular acceleration, which directly correlates with torque, effectively satisfying Hamiltonian dynamics. The unknown time constant $\alpha$ is learned as part of the control matrix $\bfg_{\bftheta}(\boldsymbol{\frakq})$ during the training process.

Given Hamiltonian network design, described in Sec. \ref{sec:learning_using_cycle_consistency}, we further specify that both the generalized mass and dissipation matrices have a block-diagonal form: $\bfM_{\bftheta}(\bffrakq) = \text{diag}(\bfM_{1,\bftheta}(\bffrakq), \bfM_{2,\bftheta}(\bffrakq)), \bfD_{\bftheta}(\bffrakq) = \text{diag}(\bfD_{\bfv,\bftheta}(\bffrakq), \bfD_{\bfomega,\bftheta}(\bffrakq))$ and only the first and last row of $\bfg_{\bftheta}(\bffrakq)$ are non-zero and to be learned from data. This aligns with the nonholonomic constraint discussed in Sec. ~\ref{subsec:ida_pbc_differential_drive}. The model was trained for 400 iterations with potential energy $\calV_{\bftheta}(\bffrakq)=0$ and a fixed nominal model with $\bar{\bfL}(\bffrakq) = \text{block-diag} \begin{bmatrix} \sqrt{m} \bfI & \bfGamma \end{bmatrix}, \bar{\bfLambda}(\bffrakq) = 0.1 \bfI, \bfg_{0}(\boldsymbol{\frakq}) = \begin{bmatrix}
\frac{\eta}{\alpha r} & 0 & 0 & 0 & 0 & -\frac{\eta w}{2 \alpha r} \\ 
\frac{\eta}{\alpha r} & 0 & 0 & 0 & 0 & \frac{\eta w}{2 \alpha r}
\end{bmatrix}^{\top}$
where the nominal mass $m = 16kg$, inertia $\bfGamma \bfGamma^{\top} = 0.4 \bfI$, wheel radius $r = 0.098m$, vehicle width $w = 0.31m$, wheel inertia $\eta =0.0458kgm^2$ and time constant $\alpha = 0.147$  \cite{Yu2009}. 

Fig. \ref{fig:real_jackal_learned_model} presents both training and testing results. The generalized mass and input gain converged similarly to the simulation results in Sec. \ref{subsec:simulation}, while the generalized inertia has off-diagonal elements due to the LiDAR's off-center position. 
The dissipation matrices have large values in $[\bfD_{\bfv, \bftheta}]_{1,1}$  and $[\bfD_{\bfomega, \bftheta}]_{1,1}$ consistent with the nonholonomic constraints, while $[\bfD_{\bfv, \bftheta}]_{0,0}$ and $[\bfD_{\bfomega, \bftheta}]_{2,2}$ show smaller friction along $x$ axis and around $z$ axis. 

Finally, we demonstrate our energy-based controller's performance using both nominal and learned models, in two experiments: (i) pose stabilization and (ii) trajectory tracking. For pose stabilization, both models successfully achieved the desired pose, as shown in Fig. \ref{fig:real_jackal_learned_model}, utilizing gains we tuned for the nominal model with $(k_{\bfp}, k_{\bfR_1}, k_{\bfR_2}, \bfK_{\bfd}) = (0.5, 0.5, 0.5, \text{block-diag}(1.5 \bfI, 0.05 \bfI))$. But in case of tracking, the learned model outperforms the nominal model as seen in Figure \ref{fig:tracking} without any additional controller tuning.

\section{Conclusions}
We developed an approach for learning robot dynamics from point-cloud observations via Hamiltonian neural ODEs, design an energy-shaping trajectory tracking controller for the learned dynamics and apply to wheeled nonholonomic robots. Future work will focus on dynamics learning from images and online adaptation to changing conditions. 

\newpage
\section*{APPENDIX: Partial Derivatives Derivations}
\label{sec:partial_derivatives_derivations}
In this appendix, we provide a detailed summary of all the partial derivatives used throughout the paper. The specific partial derivatives used in the various equations and their corresponding definitions are elaborated in this section. Beginning with the partial derivatives with respect to the position error:
\begin{equation}
    \label{eq:appendix}
    \begin{aligned}
        & \frac{\partial {\bf R}({\bf p}_e)}{\partial {\bf p}_{ei}} = \frac{{\bf R}({\bf e}_i)}{||{\bf p}_e||} -\frac{{\bf p}_{ei}}{||{\bf p}_e||^2} {\bf R}({\bf p}_e), \\
        & \frac{\partial \calV_{R_1}(\boldsymbol{\frakq}_e)}{\partial \bfp_e} = -\frac{1}{2} \begin{bmatrix}
            \text{tr} (\bfR_{\Delta_2} \bfR_e^{\top} \bfR^{*\top} \bfR_{\Delta_1}^{\top} \bfQ_1) \\ 
            \text{tr} (\bfR_{\Delta_2} \bfR_e^{\top} \bfR^{*\top} \bfR_{\Delta_1}^{\top} \bfQ_2) \\ 
            0
        \end{bmatrix}, \\
         & \frac{\partial \calV_{R_2}(\boldsymbol{\frakq}_e)}{\partial \bfp_e} = -\frac{k_{\bfR_1}}{2} \begin{bmatrix}
             \text{tr} (\bfR^{*\top} \bfR_{\Delta_1}^{\top} \bfQ_1) \\ 
             \text{tr} (\bfR^{*\top} \bfR_{\Delta_1}^{\top} \bfQ_2) \\ 
             0
         \end{bmatrix}, \\
         & \frac{\partial  \calV_{R_3}(\boldsymbol{\frakq}_e)}{\partial \bfp_e} = -\frac{k_{\bfR_2}}{2} \begin{bmatrix}
             \text{tr} (\bfR_e^{\top} \bfR^{*\top} \bfR_{\Delta_1}^{\top} \bfQ_1) \\ 
             \text{tr} (\bfR_e^{\top} \bfR^{*\top} \bfR_{\Delta_1}^{\top} \bfQ_2) \\ 
             0
         \end{bmatrix}, 
    \end{aligned}
\end{equation}
where $\bfQ_1 = \frac{\partial \bfR(\bfp_e)}{\partial \bfp_{e1}}$ and $\bfQ_2 = \frac{\partial \bfR(\bfp_e)}{\partial \bfp_{e2}}$. The vector ${\bf e}_i  \in \mathbb{R}^3$ represents the \(i^{th}\) standard basis vector. Then, the partial derivatives with respect to the rotation error: 
\begin{equation}
    \label{eq:appendix}
    \begin{aligned} 
        & \frac{\partial \calV_{R_1}(\boldsymbol{\frakq}_e)}{\partial {\bf R}_e}  = -\frac{1}{2} \bfR^{*\top} \bfR_{\Delta_1}^{\top} \bfR(\bfp_e) \bfR_{\Delta_2}, \\ 
        & \frac{\partial \calV_{R_3}(\boldsymbol{\frakq}_e)}{\partial {\bf R}_e}  = -\frac{k_{\bfR_2}}{2} \bfR^{*\top} \bfR_{\Delta_1}^{\top} \bfR(\bfp_e).
    \end{aligned}
\end{equation}





\bibliographystyle{IEEEtran}
\bibliography{bib/ref.bib, bib/thai_ref.bib}

\end{document}